%% file: main.tex
\pdfoutput=1 

\documentclass[11pt]{article}

\usepackage[]{acl}

\usepackage{times}
\usepackage{latexsym}

\usepackage[T1]{fontenc}

\usepackage[utf8]{inputenc}

\usepackage{microtype}

\usepackage{microtype}
\usepackage{amsmath,amssymb}
\usepackage{epsfig,graphicx,subfigure,caption}
\usepackage{algpseudocode}
\usepackage[normalem]{ulem}
\usepackage{amsmath}
\usepackage[linesnumbered,algoruled,boxed,noend]{algorithm2e}
\usepackage{xcolor}
\usepackage{color, colortbl}
\usepackage{multirow,booktabs, hhline}
\usepackage{amsmath, bm}
\usepackage[linesnumbered,algoruled,boxed,noend]{algorithm2e}
\usepackage{wrapfig}

\title{\textit{Good Examples Make A Faster Learner~}\\
Simple Demonstration-based Learning for Low-resource NER}

\author{
Dong-Ho Lee\textsuperscript{1},~
Akshen Kadakia\textsuperscript{1}\thanks{~~{Authors contributed equally.}},~
Kangmin Tan\textsuperscript{1}$^*$,~
Mahak Agarwal\textsuperscript{1},~
Xinyu Feng\textsuperscript{1},~\\
\textbf{Takashi Shibuya\textsuperscript{2},~
Ryosuke Mitani\textsuperscript{2},~
Toshiyuki Sekiya\textsuperscript{2},~
Jay Pujara\textsuperscript{1},~
Xiang Ren\textsuperscript{1}}
\\
\textsuperscript{1}Department of Computer Science, University of Southern California\\
\textsuperscript{2}R\&D Center, Sony Group Corporation\\
{\small \texttt{\{dongho.lee,akshenhe,kangmint,mahakaga,xinyuf,jpujara,xiangren\}@usc.edu}}\\
{\small \texttt{\{Takashi.Tak.Shibuya,Ryosuke.Mitani,Toshiyuki.Sekiya\}@sony.com}}
}

\begin{document}
\maketitle
\input{section/0_abstract}
\input{section/1_introduction}
\input{section/2_related}
\input{section/3_problem}

\input{section/4_method}
\input{section/5_experiments}
\input{section/6_results}

\input{section/7_conclusion}

\bibliographystyle{acl_natbib}
\bibliography{custom}

\newpage
\input{section/99_appendix}

\end{document}

%% file: section/0_abstract.tex
\begin{abstract}
Recent advances in prompt-based learning have shown strong results on few-shot text classification by using cloze-style templates.
Similar attempts have been made on named entity recognition (NER) which manually design templates to predict entity types for every text span in a sentence.
However, such methods may suffer from error propagation induced by entity span detection, high cost due to enumeration of all possible text spans, and omission of inter-dependencies among token labels in a sentence.
Here we present a simple demonstration-based learning method for NER, which lets the input be prefaced by task demonstrations for in-context learning.
We perform a systematic study on demonstration strategy regarding what to include (entity examples, with or without surrounding context), how to select the examples, and what templates to use. 
Results on in-domain learning and domain adaptation show that the model's performance in low-resource settings can be largely improved with a suitable demonstration strategy (e.g., 4-17\% improvement on 25 train instances). 
We also find that good demonstration can save many labeled examples and consistency in demonstration contributes to better performance.
\footnote{\href{https://github.com/INK-USC/fewNER}{https://github.com/INK-USC/fewNER}}
\end{abstract}

%% file: section/1_introduction.tex
\section{Introduction}

\begin{figure*}[t!]
\vspace{-0.4cm}
    \centering 
    \includegraphics[scale=1]{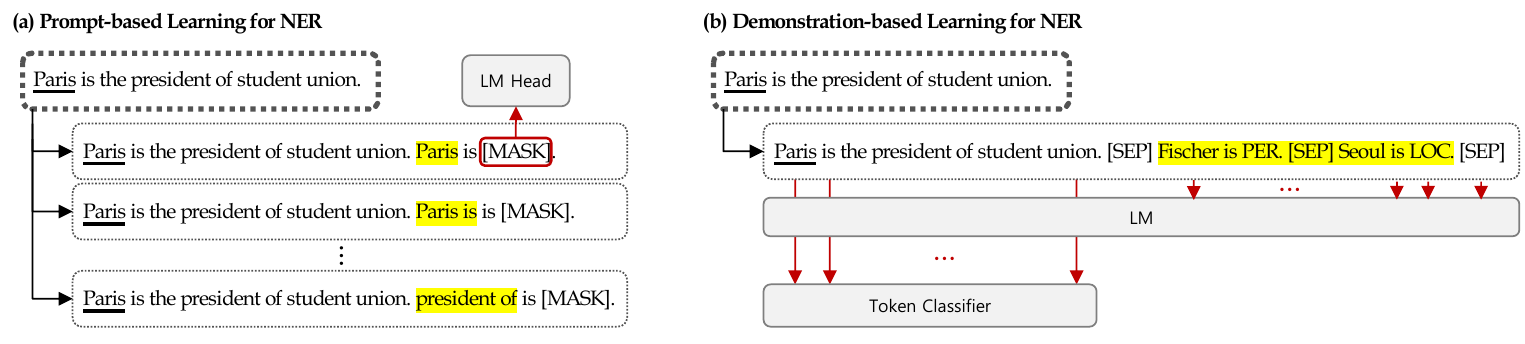}
    \caption{Prompt-based learning frameworks for NER mostly neglect entity span detection which leads to a huge time cost to generate prompts over all the entity candidates in the sentence, while our demonstration-based learning framework integrates prompt into the input itself to make better input representations for the token classification.
    }
    \label{fig:overview}
\end{figure*}

Neural sequence models have become the \textit{de facto} approach for named entity recognition (NER) and have achieved state-of-the-art results on various NER benchmarks~\citep{Lample2016NeuralAF, Ma2016EndtoendSL, liu2017empower}. However, these data-hungry models often rely on large amounts of labeled data manually annotated by human experts, which are expensive and slow to collect~\cite{huang2020few, ding-etal-2021-nerd}, especially for specialized domains (\textit{e.g.}, research papers).
To improve NER performance on low-resource (label scarcity) settings, prior works seek auxiliary supervisions, such as entity dictionary~\cite{Peng2019DistantlySN, autoner, yangner, Liu2019TowardsIN} and labeling rules~\cite{safranchik:aaai20, jiang2020cold}, to either augment human-labeled data with pseudo-labeled data, or incorporate meta information such as explanation~\cite{lin-etal-2020-triggerner,lee-etal-2020-lean,lee2021autotrigger}, context~\cite{wang-etal-2021-improving}, and prompts~\cite{ding2021prompt,cui2021template} to facilitate training. However, such methods have the following challenges:
(1) human efforts to create auxiliary supervisions (e.g., dictionaries, rules, and explanations);
(2) high computational cost to make predictions.
For example, ~\citet{ding2021prompt} shows effectiveness on entity type prediction given the entity span by constructing a prompt with the structure \textit{``[entity span] is [MASK]"}.
However, when the entity span is not given, cloze-style prompts need to be constructed over all the entity candidates in the sentence with the structure \textit{``[entity candidate] is [MASK]"} to make a prediction~\cite{cui2021template}. Such brute-force enumerations are often expensive. 


In this paper, we propose \textit{demonstration-based learning}~\cite{gao-etal-2021-making,liu2021pre}, a simple-yet-effective way to incorporate automatically constructed auxiliary supervision. The idea was originally proposed in prompt-based learning to show some task examples before the cloze-style template so that the model can better understand and predict the masked slot~\cite{gao-etal-2021-making}.
This paper proposes modified version of demonstration-based learning for NER task.
Instead of reformatting the NER task into the cloze-style template, we augment the original input instances by appending automatically created task demonstrations and feed them into pre-trained language models (PTLMs) so that the model can output improved token representations by better understandings of the tasks.
Unlike existing efforts which require additional human labor to create such auxiliary supervisions, our model can be automatically constructed by picking up proper task examples from the train data.
Moreover, unlike approaches that need to change the format of token classification into cloze-style mask-filling prediction which can neglect latent relationships among token labels, our approach can be applied to existing token classification module in a plug-and-play manner (See Figure~\ref{fig:overview} (a) vs (b)).



We investigate the effectiveness of task demonstration in two different low-resource settings: (1) in-domain setting which is a standard NER benchmark settings where the train and test dataset come from the same domain; and (2) domain-adaptation setting which uses sufficient labeled data in source domain to solve new tasks in a target domain.
Here, we study which variants of task demonstration are useful to train an accurate and label-efficient NER model and further explore ways to adapt the source model to target domain with a small amount of target data.
We propose two ways of automatic task demonstration construction: 
(1) \textit{entity-oriented demonstration} selects an entity example per entity type from train data to construct the demonstration. It allows the model to get a better sense of entity type by showing its entity example; and
(2) \textit{instance-oriented demonstration} retrieves instance example similar to input sentence in train data. It allows the model to get a better sense of the task by showing similar instances and their entities.

We show extensive experimental results on CoNLL03, Ontonotes 5.0 (generic domain), and
BC5CDR (biomedical domain) over 3 different templates and 5 selection/retrieval strategies for task demonstrations.
For \textit{entity-oriented demonstration}, we present 3 selection strategies to choose appropriate entity example per entity type:
(1) \texttt{random} randomly selects entity example per entity type;
(2) \texttt{popular} selects the entity example which occurs the most per entity type in the train data; and
(3) \texttt{search} selects the entity example per entity type that shows the best performance in the development set.
And for \textit{instance-oriented demonstration}, we present 2 retrieval strategies to choose appropriate instance example (\texttt{SBERT}~\cite{reimers-2019-sentence-bert} vs. \texttt{BERTScore}~\cite{bert-score}).

\begin{figure*}[t!]
\vspace{-0.4cm}
    \centering 
    \includegraphics[scale=1]{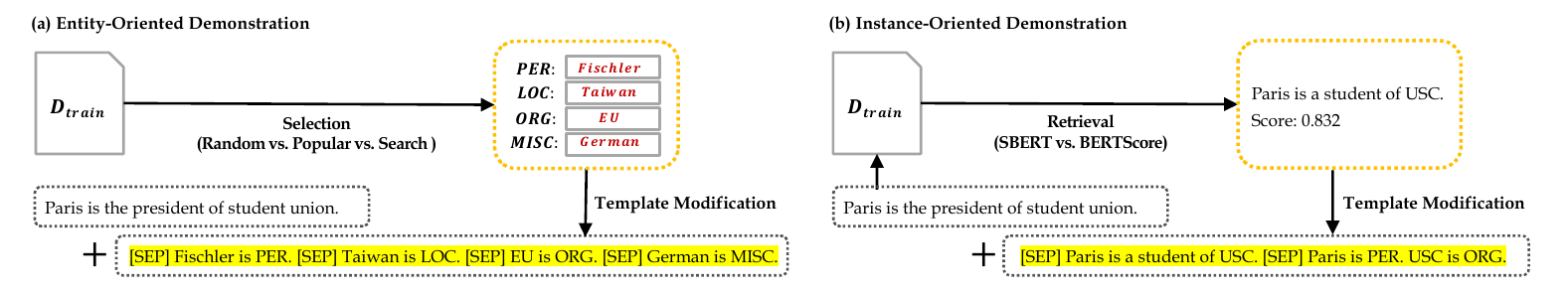}
    \caption{\textbf{Task Demonstration for NER.} (a) Entity-oriented demonstration selects an entity example per each entity type from the train data to append to the sentence; while (b) instance-oriented demonstration retrieves an instance from the train data to append to the sentence (along with the entities therein).}
    \label{fig:demonstration}
\end{figure*}

Our findings include:
(1) good demonstration can save many labeled examples to reach a similar level of performance in low-resource settings. Our approach consistently outperforms standard fine-tuning by up to 3 points in terms of F1 score (p-value < 0.02);
(2) demonstration becomes more effective when we also provide context. For example, not only showing `Fischler is PER', but also the sentence that contains `Fischler' as person, such as `France backed Fischler's proposal'; and
(3) consistency in demonstration contributes to better performance. Our experiments show that using consistent demonstration for all instances rather than varying per instance lead to better performance 

%% file: section/2_related.tex
\section{Related Works}

\paragraph{NER with additional supervision}
Recent attempts addressing label scarcity have explored various types of human-curated resources as auxiliary supervision.
One of the research lines to exploit such auxiliary supervision is distant-supervised learning.
These methods use entity dictionaries~\citep{Peng2019DistantlySN, autoner, yangner, Liu2019TowardsIN} or labeling rules~\citep{safranchik:aaai20, jiang2020cold} to generate noisy-labeled data for learning a NER model.
Although these approaches largely reduce human efforts in annotation, the cross-entropy loss may make the model be overfitted to the wrongly labeled tokens due to noisy labels~\cite{meng2021distantly}.
Another line of research is incorporating such auxiliary supervision during training and inference in a setting of supervised learning.
These approaches usually incorporate external information that is encoded including POS labels, syntactic constituents, dependency relations~\citep{syntactic_nie-etal-2020-improving, syntactic_Tian2020ImprovingBN}, explanations~\citep{lin-etal-2020-triggerner, lee-etal-2020-lean, lee2021autotrigger}, retrieved context~\cite{wang-etal-2021-improving} and prompts~\cite{ding2021prompt,cui2021template}.

\paragraph{Demonstration-based Learning}
Providing a few training examples in a natural language prompt has been widely explored in autoregressive LMs~\cite{NEURIPS2020_1457c0d6, Zhao2021Calibrate}.
Such prompt augmentation is called demonstration-based learning~\cite{gao-etal-2021-making}.
This is designed to let prompt be prefaced by a few examples before it predicts label words for \textit{[MASK]} in the cloze-style question.
Recent works on this research line explore a good selection of training examples~\cite{gao-etal-2021-making} and permutation of them as demonstration~\cite{kumar-talukdar-2021-reordering}.


%% file: section/3_problem.tex
\section{Problem Definition}

In this section, we introduce basic concepts of named entity recognition, standard fine-tuning for sequence labeling, and domain adaptation for sequence labeling.
We then formally introduce our goal -- generating task demonstration and then developing a learning framework that uses them to improve NER models.

\vspace{-0.1cm}
\subsection{Named Entity Recognition}
Here, we let $\mathbf{x} = [x^{(1)}, x^{(2)}, \dots x^{(n)}]$ denote the sentence composed of a sequence of $n$ words and $\mathbf{y} = [y^{(1)}, y^{(2)}, \dots y^{(n)}]$ denote the sequence of NER tags.
The task is to predict the entity tag $y^{(i)}\in\mathcal{Y}$ for each word $x^{(i)}$, where $\mathcal{Y}$ is a pre-defined set of tags such as \{\textsc{B-PER}, \textsc{I-PER}, \dots, \textsc{O}\}.
In \textit{standard fine-tuning}, NER model $\mathcal{M}$ parameterized by $\theta$ is trained to minimize the cross entropy loss over token representations $\mathbf{h} = [h^{(1)}, h^{(2)}, \dots h^{(n)}]$ which are generated from the pre-trained contextualized embedder as follows:
\begin{equation}
\mathcal{L}=-\sum_{i=1}^{n} \log f_{i, y_{i}}(\mathbf{h} ; \boldsymbol{\theta})
\end{equation}
where $f$ is the model's predicted conditional probability that can be either from linear or CRF layers.

\subsection{In-domain Low-resource Learning}
We let $\mathcal{D}_{\text{train}}$ and $\mathcal{D}_{\text{test}}$ denote the labeled train and test dataset, respectively, consisting of $\left\{\left(\mathbf{x}_{\mathbf{i}}, \mathbf{y}_{\mathbf{i}}\right)\right\}$.
Here, we expect the number of labeled instances in $\mathcal{D}_{\text{train}}$ is extremely limited (e.g., $N<50$).
Given such small labeled instances, our goal is to train an accurate NER model with task demonstrations compared to standard fine-tuning and show the effectiveness of demonstration-based learning.
We evaluate the trained models on $\mathcal{D}_{\text{test}}$.

\subsection{Low-resource Domain Adaption}
Domain adaptation aims to exploit the abundant data of well-studied source domains to improve the performance in target domains of interest.
We consider two different settings: (1) \textit{label-sharing} setting in which the label space $\mathbf{L}=\left\{l_{1}, \ldots, l_{|L|}\right\}$ (e.g., $l_{i}=PERSON$) of source-domain data $\mathcal{S}$ and target-domain data $\mathcal{T}$ are equal; (2) \textit{label-different} setting which $\mathbf{L}$ is different.

\begin{figure*}[t!]
\vspace{-0.4cm}
    \centering 
    \includegraphics[scale=0.9]{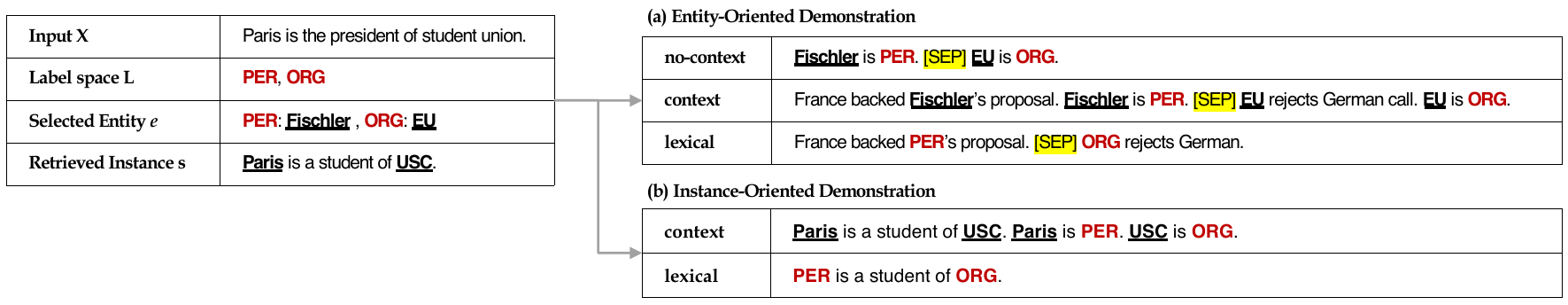}
    \vspace{-0.5cm}
	\caption{\textbf{Demonstration Template $T$}. Given input $\mathbf{x}$ and label space $\mathbf{L}$, entity-oriented demonstration selects entity $e$ per each label $l \in \mathbf{L}$ to construct three types of templates (\texttt{no-context}, \texttt{context}, \texttt{lexical}) while instance-oriented demonstration retrieve instance $s$ to create two types of templates (\texttt{context}, \texttt{lexical}).}
    \label{fig:template}
\end{figure*}

In domain adaptation, we first train a model $\mathcal{M}_{s}$ on source-domain data $\mathcal{S}$.
Next, we initialize the weights of the new model $\mathcal{M}_{t}$ by weights of $\mathcal{M}_{s}$.
Here, we can either transfer the whole model weights or only the weights of contextualized embedder from $\mathcal{M}_{s}$ to $\mathcal{M}_{t}$.
Then, we further tune $\mathcal{M}_{t}$ on target-domain data $\mathcal{T}$.
In our preliminary experiments, we find that transferring only the embedder from $\mathcal{M}_{s}$ to $\mathcal{M}_{t}$ is much more effective than transferring the whole model weights (See first rows in Table~\ref{tab:indomain} and Table~\ref{tab:crossdomain}).
For this paper, we focus on the effectiveness of our models to adapt to the target domain with a $\mathcal{T}$, for which the number of instances is extremely limited. We then compare the results of tasks with demonstration to those without demonstration.

%% file: section/4_method.tex
\section{Demonstration-based NER}
In this work, we focus on how to create effective task demonstration $\tilde{\mathbf{x}}$ to elicit better token representations for $\mathbf{x}$, and then we propose an efficient learning framework that can be improved by the effect of $[\mathbf{x} ; \tilde{\mathbf{x}}]$. 
This section introduces the concepts of \textit{demonstration-based learning}, and provides details of the approach.
Here, we study example sampling strategies and templates to construct the demonstration (Sec~\ref{ssec:demonstration}) and how we can train the NER model with the demonstration (Sec~\ref{ssec:model}).

\subsection{Task Demonstration}
\label{ssec:demonstration}
Task demonstration $\tilde{\mathbf{x}}=\left[\right.$\textit{[SEP]}$\left.; \hat{\mathbf{x}}_{1} ; \cdots ; \hat{\mathbf{x}}_{l}\right]$
is constructed by selecting entity example $e$ or retrieving instance example $s$ from $\mathcal{D}_{\text{train}}$ ($\mathcal{T}_{\text{train}}$ for domain adaptation) and modifying by template $T$ to form $\hat{\mathbf{x}}_{i}$.
The demonstration sequence $\tilde{\mathbf{x}}$ is then appended to the original input $\mathbf{x}$ to create a demonstration-augmented input $[\mathbf{x} ; \tilde{\mathbf{x}}]$.
Here, \textit{[SEP]} in front of $\tilde{\mathbf{x}}$ is to separate $\mathbf{x}$ and $\tilde{\mathbf{x}}$.
The key challenge of constructing task demonstration is to choose appropriate $e$ or $s$ and template $T$ that can be helpful to demonstrate how the model should solve the task.
As shown in Figure~\ref{fig:demonstration}, we categorize the demonstration into (1) \textit{entity-oriented demonstration}; and (2) \textit{instance-oriented demonstration} by whether we choose $e$ or $s$ respectively, for demonstration.

\paragraph{Entity-oriented demonstration.} 
Given an entity type label set $\mathbf{L}=\left\{l_{1}, \ldots, l_{|L|}\right\}$, we select an entity example $e$ per label $l$ from $\mathcal{D}_{\text{train}}$.
Then, we modify it using template $T$.
To select $e$ per each $l$, we first enumerate all the $e \in \mathcal{D}_{\text{train}}$ and create a mapping $\left\{l_{i} : \left[e_{1}, \ldots, e_{n}\right] \mid l_{i} \in \mathbf{L} \right\}$ between $l$ and corresponding list of entities.
Then for each label $l$, we select $e$ by three selection strategies:
(1) \texttt{random} randomly chooses $e$ from the list;
(2) \texttt{popular} chooses $e$ that occurs the most frequently in the list; and
(3) \texttt{search} conducts grid search over possible entity candidates per label. Here, we sample top-k frequent entities per label, and search over combinations of entity candidates ($=k^{|L|}$). 
We find the best combination that maximizes the F1 score on the dev set $\mathcal{D}_{\text{dev}}$.
Here, $\tilde{\mathbf{x}}_\mathbf{i}$ for every $\mathbf{x}_\mathbf{i}$ is different in \texttt{random} while $\tilde{\mathbf{x}}_\mathbf{i}$ for every $\mathbf{x}_\mathbf{i}$ is same in \texttt{popular} and \texttt{search}.

\paragraph{Instance-oriented demonstration.} Given an input $\mathbf{x}$, we retrieve an instance example $s$ that is the most relevant to the input from $\mathcal{D}_{\text{train}}$. Then, we modify the $s$ along with its $\left\{e,l\right\} \in s$ by template $T$.
For retrieval, we present two strategies: 
(1) \texttt{SBERT}~\cite{reimers-2019-sentence-bert} retrieves semantically similar sentence using pre-trained bi-encoder. It produces CLS embeddings independently for an input $\mathbf{x}$ and $s \in \mathcal{D}_{\text{train}}$, and compute the cosine similarity between them to rank $s \in \mathcal{D}_{\text{train}}$;
(2) \texttt{BERTScore}~\cite{bert-score}, which is originally used as a text generation metric, retrieves token-level semantically similar sentence by computing a sum of cosine similarity between token representations of two sentences.
Since the NER task aims to token classification, sentence-level similarity may retrieve a sentence that is semantically relevant but has no relevant entities.

\paragraph{Fixed vs Variable demonstration.}
As described in previous sections, the demonstration in some strategies varies per instance while in others it stays fixed globally. We can divide the demonstration strategies into two categories:
(1) Variable demonstration: \texttt{random}, \texttt{SBERT}, \texttt{BERTScore}
(2) Fixed demonstration: \texttt{popular}, \texttt{search}

\paragraph{Demonstration template.}
As shown in Figure~\ref{fig:template}, we select three variants of template $T$:

\noindent
(1) \texttt{no-context} shows selected $e$ per $l$ with a simple template \text{``$e$ \textit{is} $l$."}, without including the specific sentence where the entities show up. Between each pair of $(e,l)$ (of different entity labels $l$), we concatenate with separator \textit{[SEP]}. This template is only applied to the entity-oriented demonstration.

\noindent
(2) \texttt{context} in entity-oriented demonstration shows selected $e$ per $l$ along with an instance sentence $s$ that contains $e$ as a type of $l$. For each triple of $(e,l,s)$, it is modified into \text{``$s$. $e$ \textit{is} $l$."} and concatenated with \textit{[SEP]}. 
For instance-oriented demonstration, it shows the retrieved instance $s$ along with all the entities mentioned in the sentence $e \in s$. It is modified into \text{``$s$. $e_{1}$ \textit{is} $l_{1}$. \ldots $e_{n}$ \textit{is} $l_{n}$."}.

\noindent
(3) \texttt{lexical} in entity-oriented demonstration also shows selected $e$ per $l$ along with an instance sentence $s$.
But here we only show $s$, which the entity span $e$ is replaced by its label string $l$.
For instance-oriented demonstration, we show retrieved $s$ by replacing $e \in s$ with the corresponding $l$.
We expect such templates can form labeling rules and let the model know how to label the sentence.

\subsection{Model Training with Demonstration}
\label{ssec:model}
Transformer-based standard fine-tuning for NER first feeds the input sentence $\mathbf{x}$ into a transfomer-based PTLMs to get the token representations $\mathbf{h}$.
The token representations $\mathbf{h}$ are fed into a CRF layer to get the conditional probability $p_{\theta}(\mathbf{y} \mid \mathbf{h})$, and the model is trained by minimizing the conditional probability by cross entropy loss:
\begin{equation}
\mathcal{L}=-\sum_{i=1}^{n} \log p_{\theta}(\mathbf{y} \mid \mathbf{h})
\label{eq:ce}
\end{equation}

In our approach, we define a neural network parameterized by $\theta$ that learns from a concatenated input $[\mathbf{x} ; \tilde{\mathbf{x}}]$.
For both model training and inference, we feed the input and retrieve the representations:
\begin{equation}
\scalebox{.8}{$
[\mathbf{h} ; \tilde{\mathbf{h}}] = [h^{(1)}, \dots h^{(n)}, \tilde{h}^{(1)}, \dots \tilde{h}^{(n)}]=\operatorname{embed}([\mathbf{x} ; \tilde{\mathbf{x}}])$
}
\end{equation}
As shown in Figure~\ref{fig:overview}, we then feed $\mathbf{h}$ into the CRF layer to get predictions and train by minimizing the conditional probability $p_{\theta}(\mathbf{y} \mid \mathbf{h})$ as Equation~\ref{eq:ce}.

For domain adaptation, we first train $\mathcal{M}_{s}$ with standard fine-tuning. Then, transfer the weights of embedder of $\mathcal{M}_{s}$ to $\mathcal{M}_{t}$ and further fine-tune $\mathcal{M}_{t}$ with our approach.

%% file: section/5_experiments.tex
\begin{table}[t]
	\centering
	\scalebox{0.67}{
		\begin{tabular}{llcc}
            \toprule
            \multirow{2}{*}{\textbf{Dataset}} & \multirow{2}{*}{\textbf{Label}} & \multicolumn{2}{c}{\textbf{Train Data}} \\
            \cmidrule(lr){3-4} & & 25 & 50 \\
            \midrule
            CoNLL03 & PER (Person) & 16.0\textsubscript{$\pm$3.52} & 29.2\textsubscript{$\pm$4.52} \\
             & LOC (Location) & 15.6\textsubscript{$\pm$3.92} & 30.4\textsubscript{$\pm$4.07} \\
             & ORG (Organization) & 21.8\textsubscript{$\pm$2.31} & 32.6\textsubscript{$\pm$3.77} \\
             & MISC (Miscellaneous) & 11.0\textsubscript{$\pm$2.52} & 15.6\textsubscript{$\pm$2.33} \\
            \midrule
            Ontonotes 5.0 & PER (Person) & 10.8\textsubscript{$\pm$2.22} & 21.4\textsubscript{$\pm$4.02}  \\
             & LOC (Location) & 16.0\textsubscript{$\pm$3.52} & 25.0\textsubscript{$\pm$7.32} \\
             & ORG (Organization) & 13.8\textsubscript{$\pm$3.48} & 24.2\textsubscript{$\pm$6.17} \\
             & MISC (Miscellaneous) & 23.8\textsubscript{$\pm$5.56} & 62.6\textsubscript{$\pm$7.93} \\
            \midrule
            BC5CDR & Disease & 25.8\textsubscript{$\pm$6.01} & 29.2\textsubscript{$\pm$4.52} \\
             & Chemical & 51.0\textsubscript{$\pm$7.49} & 65.8\textsubscript{$\pm$7.12}\\
            \midrule
        \end{tabular}
	}
	\vspace{-0.2cm}
	\caption{\textbf{Data statistics.} Average number of entities per each entity type over 5 different subsamples.}
	\label{tab:statistics}
\end{table}

\begin{table*}[t]
\vspace{-0.2cm}
	\centering
	\scalebox{0.66}{
		\begin{tabular}{lcccccccc}
            \toprule
            \multirow{2}{*}{\textbf{Demonstration / Method}} & \multirow{2}{*}{\textbf{Strategy}} & 
            \multirow{2}{*}{\textbf{Template}} & \multicolumn{2}{c}{\textbf{CoNLL03}} & \multicolumn{2}{c}{\textbf{Ontonotes 5.0}} & \multicolumn{2}{c}{\textbf{BC5CDR}} \\
            \cmidrule(lr){4-5} \cmidrule(lr){6-7} \cmidrule(lr){8-9} & & &
            25 & 50 & 25 & 50 & 25 & 50 \\
            \midrule
            BERT+CRF w/o demonstration & - & - & 52.72 \textsubscript{$\pm$2.44} & 62.75 \textsubscript{$\pm$0.98} & 38.97 \textsubscript{$\pm$4.62} & 54.51 \textsubscript{$\pm$3.27} & 52.56 \textsubscript{$\pm$0.46} & 60.20 \textsubscript{$\pm$2.01} \\
            \midrule
            BERT+CRF w/
            & \texttt{SBERT} & \texttt{lexical} & 48.92 \textsubscript{$\pm$2.81} & 57.68 \textsubscript{$\pm$0.37} & 36.58 \textsubscript{$\pm$4.61} & 44.47 \textsubscript{$\pm$2.58} & 49.41 \textsubscript{$\pm$0.94} & 51.98 \textsubscript{$\pm$2.14} \\
            Instance-oriented demonstration & (\texttt{variable}) & \texttt{context} & 53.62 \textsubscript{$\pm$1.64} & 64.21 \textsubscript{$\pm$1.87} & 42.18 \textsubscript{$\pm$5.21} & 53.07 \textsubscript{$\pm$3.46} & 54.71 \textsubscript{$\pm$2.09} & 59.78 \textsubscript{$\pm$1.47} \\
            \cmidrule(lr){2-9}
            & \texttt{BERTScore} & \texttt{lexical} & 49.55 \textsubscript{ $\pm$3.18} & 58.85 \textsubscript{ $\pm$1.06} & 35.42 \textsubscript{ $\pm$3.88} & 44.70 \textsubscript{ $\pm$2.41} & 49.37 \textsubscript{ $\pm$0.19} & 51.61 \textsubscript{ $\pm$2.45} \\
            & (\texttt{variable}) & \texttt{context} & 53.97 \textsubscript{ $\pm$1.52} & 64.66 \textsubscript{ $\pm$2.04} & 37.56 \textsubscript{ $\pm$5.29} & 53.13 \textsubscript{ $\pm$3.22} & \underline{54.81 \textsubscript{ $\pm$2.11}} & 59.63 \textsubscript{ $\pm$1.94} \\
            \midrule
            BERT+CRF w/ 
            & \texttt{random} & \texttt{no-context} & 53.95 \textsubscript{$\pm$1.89} & 63.31 \textsubscript{$\pm$2.14} & 42.25 \textsubscript{$\pm$3.61} & 55.71 \textsubscript{$\pm$3.82} & 53.58 \textsubscript{$\pm$0.48} & 59.97 \textsubscript{$\pm$1.89}  \\
            Entity-oriented demonstration & (\texttt{variable}) & \texttt{lexical} & 55.20 \textsubscript{$\pm$2.24} & 63.60 \textsubscript{$\pm$2.32} & 44.02 \textsubscript{$\pm$4.73} & 56.31 \textsubscript{$\pm$3.83} & 53.79 \textsubscript{$\pm$0.61} & 59.65 \textsubscript{$\pm$1.71} \\
            & & \texttt{context} & 54.84 \textsubscript{$\pm$2.12} & 63.51 \textsubscript{$\pm$2.83} & 43.57 \textsubscript{$\pm$3.73} & 56.76 \textsubscript{$\pm$3.69} & 54.08 \textsubscript{$\pm$0.97} & 59.94 \textsubscript{$\pm$1.70} \\
            \cmidrule(lr){2-9}
            & \texttt{popular} & \texttt{no-context} & 54.34 \textsubscript{$\pm$3.33} & 64.30 \textsubscript{$\pm$2.76} & 43.02 \textsubscript{$\pm$4.33} & 56.65 \textsubscript{$\pm$3.35} & 53.86 \textsubscript{$\pm$0.86} & 60.51 \textsubscript{$\pm$1.77} \\
            & (\texttt{fixed}) & \texttt{lexical} & 56.22 \textsubscript{$\pm$3.88} & \underline{64.95 \textsubscript{$\pm$2.04}} & 45.31 \textsubscript{$\pm$5.02} & 58.24 \textsubscript{$\pm$3.17} & 54.14 \textsubscript{$\pm$0.67} & 60.67 \textsubscript{$\pm$1.58} \\
            & & \texttt{context} & 56.52 \textsubscript{$\pm$3.34} & 64.47 \textsubscript{$\pm$2.35} & \underline{45.52 \textsubscript{$\pm$4.69}} & 58.40 \textsubscript{$\pm$3.24} & 54.31 \textsubscript{$\pm$0.80} & \underline{61.31 \textsubscript{$\pm$1.51}} \\
            \cmidrule(lr){2-9}
            & \texttt{search} & \texttt{no-context} & 54.63 \textsubscript{$\pm$2.12} & 64.50 \textsubscript{$\pm$2.76} & 42.88 \textsubscript{$\pm$5.41} & 56.96 \textsubscript{$\pm$4.09} & 53.97 \textsubscript{$\pm$1.32} & 60.84 \textsubscript{$\pm$2.14}  \\
            & (\texttt{fixed}) & \texttt{lexical} & \underline{56.57 \textsubscript{ $\pm$3.61}} & \bf 65.11 \textsubscript{ $\pm$2.71} & 44.87 \textsubscript{ $\pm$5.09} & \underline{58.51 \textsubscript{ $\pm$3.42}} & 54.39 \textsubscript{ $\pm$1.57} & 60.76 \textsubscript{ $\pm$2.12} \\
            & & \texttt{context} & \bf 57.00 \textsubscript{ $\pm$4.03} & 64.82 \textsubscript{ $\pm$3.16} & \bf 45.74 \textsubscript{ $\pm$5.57} & \bf 59.00 \textsubscript{ $\pm$3.27} & \bf 55.83 \textsubscript{ $\pm$1.25} & \bf 62.87 \textsubscript{ $\pm$2.41} \\
            \midrule
        \end{tabular}
	}
	\vspace{-0.1cm}
	\caption{\small \textbf{In-domain performance comparison (F1-score)} on CoNLL03, Ontonotes 5.0, and BC5CDR by different number of training instances. We randomly sample $k$ training instances with a constraint that sampled instances should cover all the IOBES labels in the whole dataset. Best variants are \textbf{bold} and second best ones are \underline{underlined}.
	Scores are average of 15 runs (5 different sub-samples and 3 random seeds) and the backbone LM model is \texttt{bert-base-cased}.
	}
	\label{tab:indomain}
\end{table*}

\section{Experimental Setup}

\vspace{-0.1cm}
\subsection{Datasets}
We consider three NER datasets as target tasks.
We consider two datasets for a general domain (\textbf{CoNLL03}~\citep{conll}, \textbf{Ontonotes 5.0}~\citep{weischedel2013ontonotes}) and one dataset for a bio-medical domain (\textbf{BC5CDR}~\citep{bc5cdr}).
\textbf{CoNLL03} is a general domain NER dataset that has 22K sentences containing four types of general named entities: \textsc{location}, \textsc{person}, \textsc{organization}, and \textsc{miscellaneous} entities that do not belong in any of the three categories.
\textbf{Ontonotes 5.0} is a corpus that has roughly 1.7M words along with integrated annotations of multiple layers of syntactic, semantic, and discourse in the text.
Named entities in this corpus were tagged with a set of general 18 well-defined proper named entity types.
We split the data following ~\cite{pradhan-etal-2013-towards}.
\textbf{BC5CDR} has 1,500 articles containing 15,935 \textsc{Chemical} and 12,852 \textsc{Disease} mentions.

\begin{table}[t]
	\centering
	\small
	\resizebox{0.48\textwidth}{!}{
		\begin{tabular}{lcccccc}
            \toprule
            \multirow{3}{*}{\textbf{Baselines}} &  & 
            \multicolumn{2}{c}{\textit{Label Sharing}} &
            \multicolumn{2}{c}{\textit{Label Different}} \\
            \cmidrule(lr){3-4} \cmidrule(lr){5-6} & & \multicolumn{2}{c}{\textbf{CoNLL03 -> Ontonotes}} & \multicolumn{2}{c}{\textbf{CoNLL03 -> BC5CDR}} \\
            \cmidrule(lr){3-4} \cmidrule(lr){5-6} & &
            25 & 50 & 25 & 50 \\
            \midrule
            \multicolumn{2}{l}{BERT+CRF w/o demonstration} & 61.22 \textsubscript{$\pm$1.93} & 66.44 \textsubscript{$\pm$1.75} & 52.31 \textsubscript{$\pm$1.02} & 62.10 \textsubscript{$\pm$1.01} \\
            \multicolumn{2}{l}{NNShot} & 46.67 \textsubscript{ $\pm$5.48} & 46.34 \textsubscript{ $\pm$2.66} & 44.93 \textsubscript{ $\pm$1.78} & 48.12 \textsubscript{ $\pm$2.72} \\
            \multicolumn{2}{l}{StructShot} & 43.61 \textsubscript{ $\pm$4.58} & 43.02 \textsubscript{ $\pm$3.19} & 25.86 \textsubscript{ $\pm$4.14} & 27.81 \textsubscript{ $\pm$2.10} \\
            \midrule
        \end{tabular}
	}
	\resizebox{0.48\textwidth}{!}{
		\begin{tabular}{lcccccc}
            \midrule
                \textbf{Strategy} & \textbf{Template} & \\ 
            \midrule
            \texttt{SBERT} & \texttt{lexical} & \bf 63.34 \textsubscript{ $\pm$1.53} & 68.52 \textsubscript{ $\pm$0.98} & 53.50 \textsubscript{ $\pm$2.26} & 60.52 \textsubscript{ $\pm$0.71} \\
            (\texttt{variable}) & \texttt{context} & 62.33 \textsubscript{ $\pm$1.63} & 67.86 \textsubscript{ $\pm$0.89} & 51.93 \textsubscript{ $\pm$1.96} & 60.09 \textsubscript{ $\pm$1.27}\\
            \midrule
            \texttt{BERTScore} & \texttt{lexical} & 62.26 \textsubscript{ $\pm$1.43} & 68.68 \textsubscript{ $\pm$0.25} & 52.07 \textsubscript{ $\pm$2.11} & 59.90 \textsubscript{ $\pm$0.05} \\
            (\texttt{variable}) & \texttt{context} & 62.46 \textsubscript{ $\pm$1.69} & 67.46 \textsubscript{ $\pm$0.79} & 53.58 \textsubscript{ $\pm$1.98} & 58.95 \textsubscript{ $\pm$0.38} \\
            \midrule
            \texttt{random} & \texttt{no-context} & 62.28 \textsubscript{$\pm$1.70} & 69.32 \textsubscript{$\pm$1.34} & 53.61 \textsubscript{$\pm$1.04} & 62.57 \textsubscript{$\pm$0.97}  \\
            (\texttt{variable}) & \texttt{lexical} & 62.41 \textsubscript{$\pm$1.85} & 68.84 \textsubscript{$\pm$1.78} & 53.85 \textsubscript{$\pm$1.12} & 62.30 \textsubscript{$\pm$0.75} \\
            & \texttt{context} & 62.58 \textsubscript{$\pm$2.20} & 69.26 \textsubscript{$\pm$1.51} & 54.05 \textsubscript{$\pm$0.63} & 63.04 \textsubscript{$\pm$0.31} \\
            \midrule
            \texttt{popular} & \texttt{no-context} & 62.31 \textsubscript{$\pm$1.60} & 69.39 \textsubscript{$\pm$1.59} & 54.33 \textsubscript{$\pm$0.80} & 62.87 \textsubscript{$\pm$0.23} \\
            (\texttt{fixed}) & \texttt{lexical} & 62.50 \textsubscript{$\pm$2.41} & 69.34 \textsubscript{$\pm$1.38} & 54.30 \textsubscript{$\pm$1.12} & 63.05 \textsubscript{$\pm$0.45} \\
            & \texttt{context} & 62.59 \textsubscript{$\pm$2.38} & \underline{69.91 \textsubscript{$\pm$1.24}} & 54.45 \textsubscript{$\pm$0.96} & \underline{63.40 \textsubscript{$\pm$0.33}} \\
            \midrule
            \texttt{search} & \texttt{no-context} & 62.38 \textsubscript{ $\pm$2.47} & 69.57 \textsubscript{ $\pm$1.50} & 54.51 \textsubscript{ $\pm$2.25} & 62.93 \textsubscript{ $\pm$1.96}  \\
            (\texttt{fixed}) & \texttt{lexical} & 62.51 \textsubscript{ $\pm$2.43} & 68.93 \textsubscript{ $\pm$1.69} & \underline{54.70 \textsubscript{ $\pm$2.26}} & 62.88 \textsubscript{ $\pm$2.90} \\
            & \texttt{context} & \underline{62.63 \textsubscript{ $\pm$2.94}} & \bf 69.98 \textsubscript{ $\pm$1.63} & \bf 54.97 \textsubscript{ $\pm$1.99} & \bf63.55 \textsubscript{ $\pm$1.58} \\
            \midrule
        \end{tabular}
	}
	\vspace{-0.1cm}
	\caption{\small \textbf{Domain adaptation performance comparison (F1-score)} on Ontonotes 5.0 and BC5CDR by different number of training instances. $\mathcal{M}_{s}$ is trained on CoNLL03 and $\mathcal{M}_{t}$ is initialized with embedder of $\mathcal{M}_{s}$.
    Scores are average of 15 runs (5 different sub-samples and 3 random seeds) and the backbone LM model is \texttt{bert-base-cased}.
	}
	\label{tab:crossdomain}
	\vspace{-0.3cm}
\end{table}

\subsection{Baselines}
To show its effectiveness in few-shot NER, we also show baselines of few-shot NER methods NNShot and StructShot ~\cite{yang-katiyar-2020-simple}.
NNshot is simple token-level nearest neighbor classification system while StructShot extends NNshot with a decoding process using abstract tag transition distribution. 
Here, both the classification model and the transition distribution should be pre-trained on the source dataset.
Thus, we consider this as domain adaptation setting.

\subsection{Experiments and Implementation Details}
We implement all the baselines and our frameworks using PyTorch~\citep{NEURIPS2019_9015} and HuggingFace~\citep{wolf-etal-2020-transformers}.
We set the batch size and learning rate to 4 and 2e-5, respectively, and use \texttt{bert-base-cased} model for all the experiments.
For each variant, we run 50 epochs over 5 different sub-samples and 3 random seeds with early-stopping 20 and show its average and standard deviation of F1 scores.
Unlike existing sampling methods for few-shot NER~\cite{yang-katiyar-2020-simple}, in which the training sample refers to one entity span in a sentence, we consider a real-world setting that humans annotate a sentence.
We sub-sample data-points by random sampling with a constraint that sampled instances should cover all the BIOES labels~\cite{chiu-nichols-2016-named} in the whole dataset.
For Ontonotes, we aggregate all other entity types rather than person, location, and organization into miscellaneous to set the \textit{label sharing} setting for domain adaptation experiments.
Table~\ref{tab:statistics} presents statistics of average number of entities per entity type over 5 different sub-samples.

%% file: section/6_results.tex
\section{Experimental Results}
\vspace{-0.1cm}

We first compare the overall performance of all baseline models and our proposed framework with the amount of training data 25 and 50 to show the impact of our approach in a low-resource scenario, assuming a task that needs to be annotated from  scratch.
Then, we show performance analysis to show the effectiveness of our approach and whether the model really learns from the demonstration.

\subsection{Performance Comparison}
\paragraph{In-domain setting}
In Table~\ref{tab:indomain}, we can observe that most variants of demonstration-based learning consistently and significantly (with p-value < 0.02) outperform the baseline by a margin ranging from 1.5 to 7 F1 score in three low-resource NER datasets (25, 50 train instances respectively). It demonstrates the potential of our approach for serving as a plug-and-play method for NER models.

\paragraph{Domain adaptation setting}
First, we observe that simple domain adaptation technique can improve the performance (First rows of Table~\ref{tab:indomain} vs. Table~\ref{tab:crossdomain}).
Here, we only transfer the embedder weights of $\mathcal{M}_{s}$ to $\mathcal{M}_{t}$, and we expect the performance gain can be attributed to the embedder of $\mathcal{M}_{s}$, which is trained in task adaptive pre-training manner on NER task formats~\cite{dontstoppretraining2020}.
In Table~\ref{tab:crossdomain}, we can see that the most variants of demonstration-based learning allow the source model $\mathcal{M}_{s}$ to be adapted to the target domain in fast with a small amount of target data $\mathcal{T}$, compared to baselines without demonstration including few-shot NER methods.


\subsection{Performance Analysis}
\paragraph{Entity vs. Instance-oriented demonstration.}
\textit{instance-oriented demonstration} performs worse than \textit{entity-oriented demonstration} due to the difficulty of finding an appropriate similar instance in a low resource train data. In our analysis, we find that the average cosine similarity between retrieved example $s$ and input $x$ is less than 0.4 which shows many of the retrieved examples are not appropriate similar examples to the input.

\begin{figure}[t]
\vspace{-0.4cm}
\hspace{0.1cm}
\begin{minipage}{0.46\textwidth}
  \centerline{\includegraphics[width=7cm]{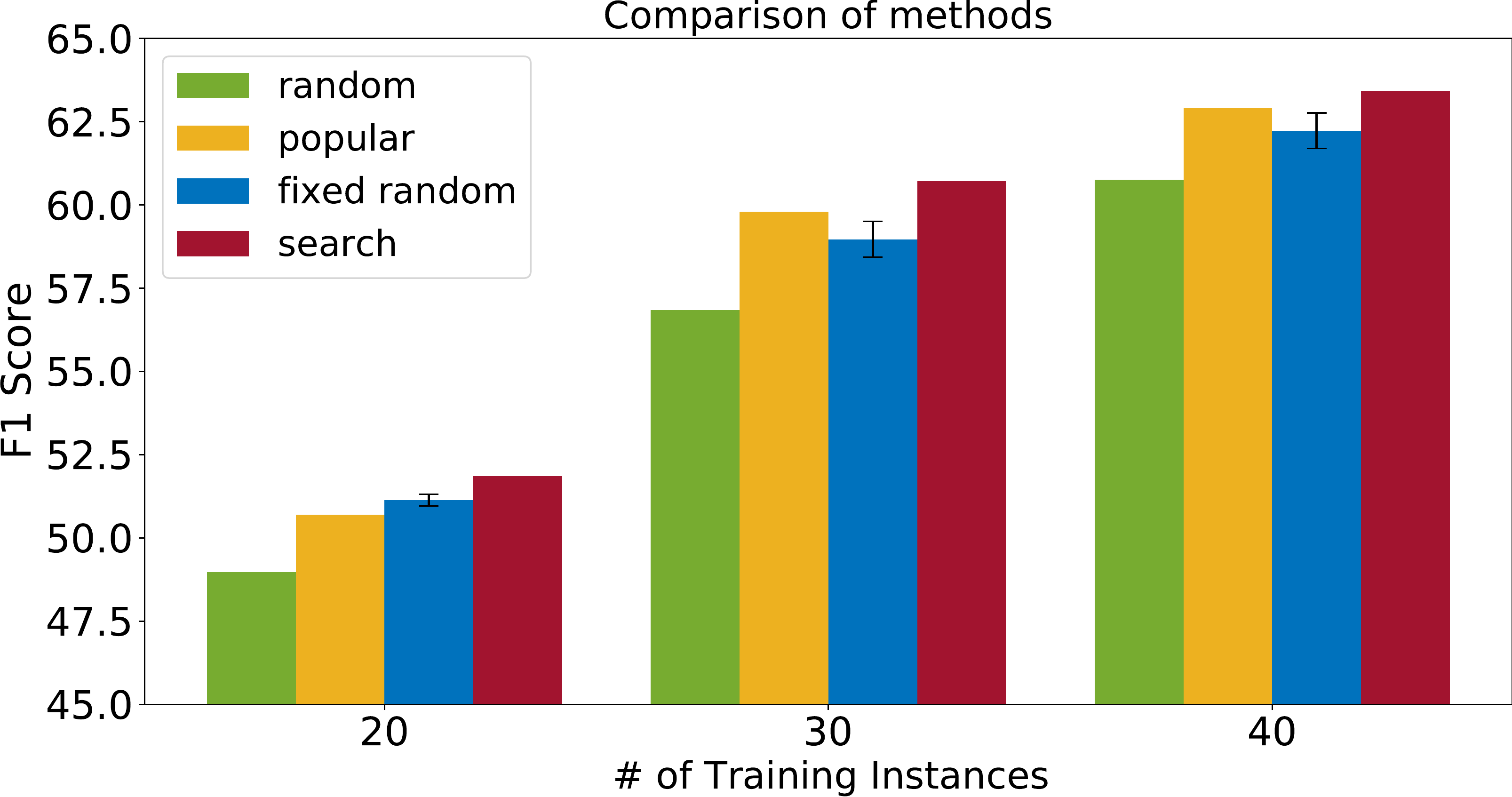}}
\end{minipage}
  \vspace{-0.2cm}
\caption{\small \textbf{Performance (F1-score) of randomly select one fixed entity per entity type for demonstration} (\texttt{fixed random}) on CoNLL03 by different numbers of train data (20, 30, 40). Error bars show standard deviation across 3 trials using 3 different random seeds for entity selection.
}
\label{fig:fixvariable}
\vspace{-0.2cm}
\end{figure}

\paragraph{Fixed vs. Variable demonstration.}
As mentioned in section 4.1, \texttt{random} doesn't pick a fixed set of demonstrations the same way as \texttt{popular} and \texttt{search}. Instead, it picks random demonstrations for each input instance. In a low-resource setting, there are often no significantly popular entities. Therefore, the fact that \texttt{popular} outperforms \texttt{random} in our experiments might suggest that the consistency of demonstration selection, rather than popularity of selected entities, is a crucial factor in better few-shot learning. To test this, we randomly select one entity per entity type and attach it as the demonstration to all instances, we call it (\texttt{fixed random}). As shown in Figure ~\ref{fig:fixvariable}, it outperforms \texttt{random} and is on par with \texttt{popular} and \texttt{search}. We believe this serves as evidence for two hypotheses: (1) consistency of demonstration is essential to performance, and (2) in low-resource settings, the effectiveness of combinations of entities as demonstrations might be a rather random function and not too affected by the combination's collective popularity in the training dataset, which further implies that the idea of \texttt{search} is on the right track.  

\paragraph{Performance in other model variants}
To show the effectiveness of demonstration-based learning as plug-and-play method, we present performance in other model variants: \texttt{bert-large-cased}, \texttt{roberta-base} and \texttt{roberta-large}.
As shown in Table~\ref{tab:modelvariant}, our method shows consistent improvement over baselines (p-value < 0.05).
It shows that demonstration-based learning can be applied to any other model variants and output better contextualized representations for NER tasks and show its potential for scalability.

\begin{table}[!t]
\vspace{-0.3cm}
	\centering
	\small
	\resizebox{0.48\textwidth}{!}{
		\begin{tabular}{ccccccc}
            \toprule
            \multirow{3}{*}{\textbf{LM}} & 
            \multirow{3}{*}{\textbf{Strategy}} & 
            \multirow{3}{*}{\textbf{Template}} & 
            \multicolumn{2}{c}{\textit{In-domain}} &
            \multicolumn{2}{c}{\textit{Label Sharing}} \\
            \cmidrule(lr){4-5} \cmidrule(lr){6-7} & & & \multicolumn{2}{c}{\textbf{CoNLL03}} & \multicolumn{2}{c}{\textbf{CoNLL03 -> Ontonotes}} \\
            \cmidrule(lr){4-5} \cmidrule(lr){6-7} & & &
            25 & 50 & 25 & 50 \\
            \midrule
            \texttt{BL} & - & - & 52.08 \textsubscript{$\pm$2.02} & 66.42 \textsubscript{$\pm$2.14} & 63.50 \textsubscript{$\pm$0.96} & 70.59 \textsubscript{$\pm$1.16} \\
            \texttt{RB} & - & - & 59.67 \textsubscript{$\pm$4.65} & 70.17 \textsubscript{$\pm$3.93} & 68.43 \textsubscript{$\pm$2.09} & 74.11 \textsubscript{$\pm$1.19} \\
            \texttt{RL} & - & - & 59.15 \textsubscript{$\pm$2.93} & 71.51 \textsubscript{$\pm$3.44} & 68.16 \textsubscript{$\pm$2.65} & 74.45 \textsubscript{$\pm$1.02}\\
            \midrule
            \texttt{BL} & \texttt{popular} & \texttt{context} & 57.60 \textsubscript{$\pm$3.37} & 67.11 \textsubscript{$\pm$2.31} & 64.09 \textsubscript{$\pm$2.95} & 70.88 \textsubscript{$\pm$1.09}\\
            \texttt{RB} & \texttt{popular} & \texttt{context} & 59.76 \textsubscript{$\pm$4.27} & 70.21 \textsubscript{$\pm$3.41} & 69.09 \textsubscript{$\pm$2.63} & 74.53 \textsubscript{$\pm$1.32} \\
            \texttt{RL} & \texttt{popular} & \texttt{context} & 59.99 \textsubscript{$\pm$2.16} & 72.15 \textsubscript{$\pm$3.81} & 68.78 \textsubscript{$\pm$2.89} & 74.93 \textsubscript{$\pm$1.07} \\
            \midrule
        \end{tabular}
	}
	\vspace{-0.2cm}
	\caption{\small \textbf{Performance comparison (F1-score)} with various backbone LMs: \texttt{bert-large-cased (BL)}; \texttt{roberta-base (RB)}; and \texttt{roberta-large (RL)}.
    Scores are average of 15 runs (5 different sub-samples and 3 random seeds).
	}
	\label{tab:modelvariant}
\end{table}

\begin{figure}[t]
\vspace{-0.2cm}
\hspace{0.1cm}
\begin{minipage}{0.23\textwidth}
  \centerline{\includegraphics[width=4.1cm]{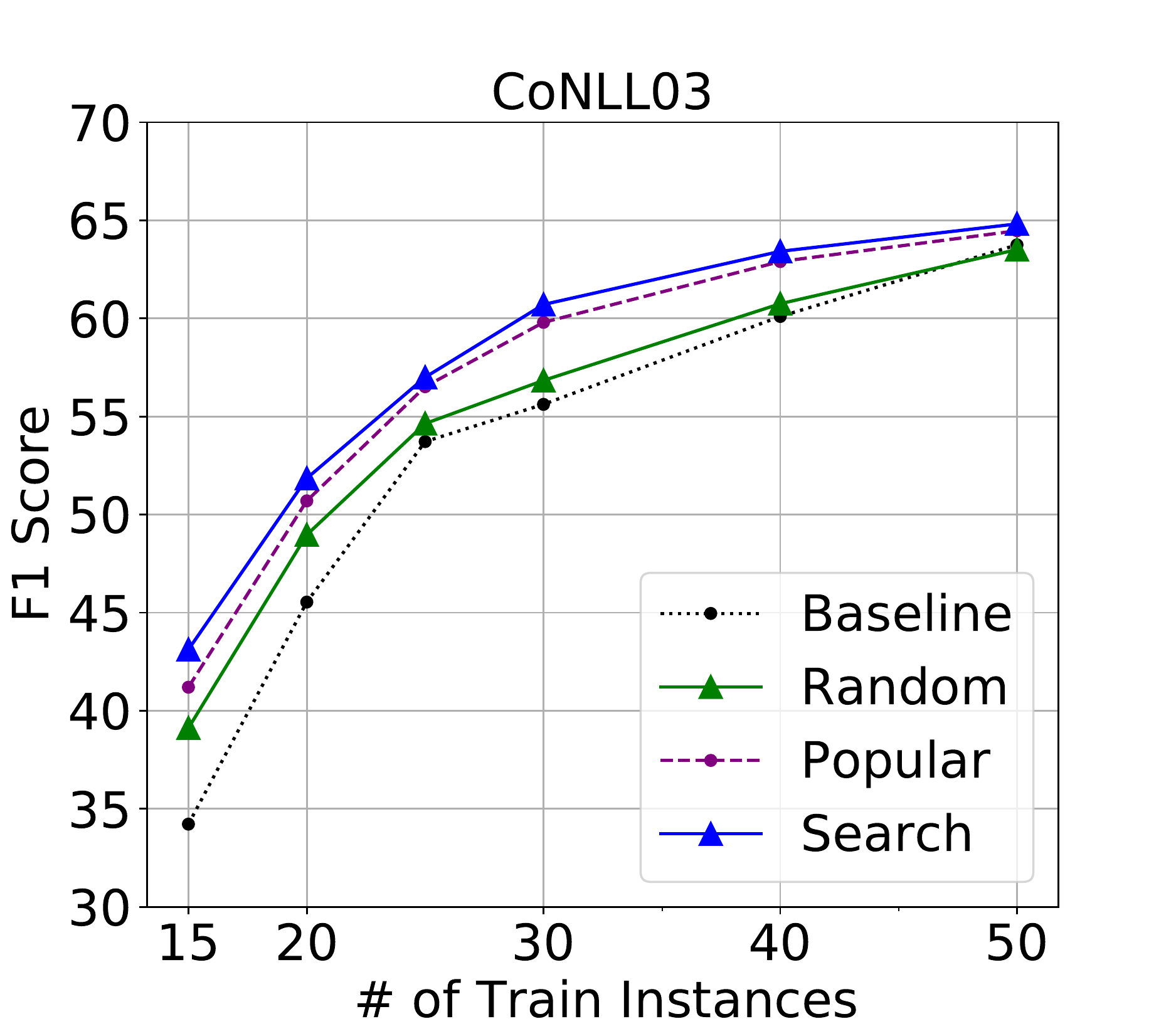}}
  \vspace{-0.1cm}
  \centerline{\small (a) CoNLL03}
\end{minipage}
\begin{minipage}{0.23\textwidth}
  \centerline{\includegraphics[width=4.1cm]{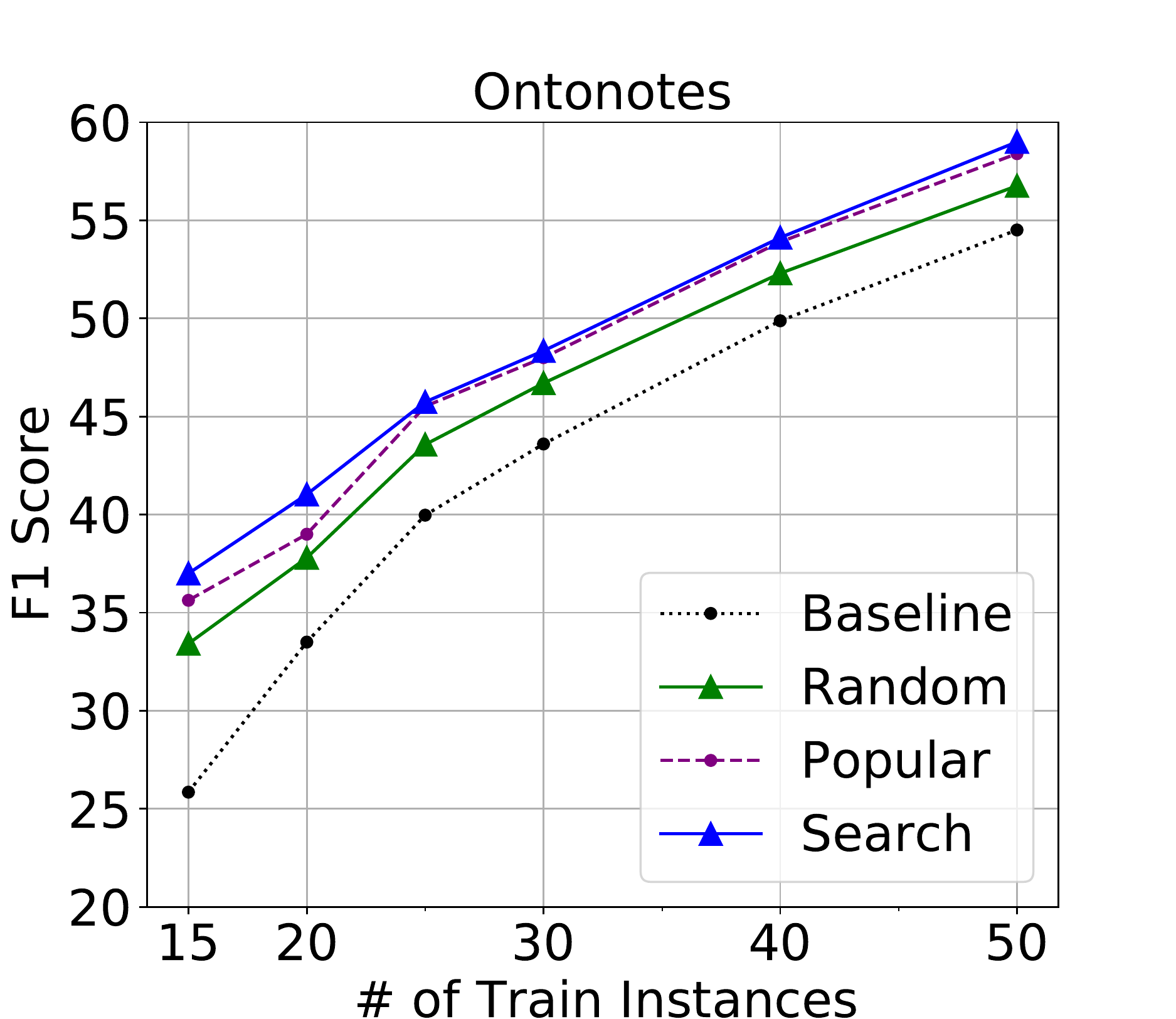}}
  \vspace{-0.1cm}
  \centerline{\small (b) Ontonotes}
\end{minipage}
  \vspace{-0.2cm}
\caption{\small \textbf{Performance (F1-score) trend with entity-oriented demonstration} on CoNLL03 and Ontonotes by different numbers of train data (15, 20, 30, 40, 50).
}
\label{fig:trend}
\vspace{-0.1cm}
\end{figure}

\paragraph{Effectiveness of \texttt{search}.}
\texttt{search} consistently outperforms all other strategies. It shows that not only the entity selection, but also the combination of entity examples per each entity type affects the performance.
To see whether it consistently outperforms the baseline over various low-resource data points, we show the performance trend of \textit{entity-oriented demonstration} in Figure~\ref{fig:trend}.

\begin{figure}[t]
\vspace{-0.4cm}
\hspace{0.1cm}
\begin{minipage}{0.23\textwidth}
  \centerline{\includegraphics[width=4.2cm]{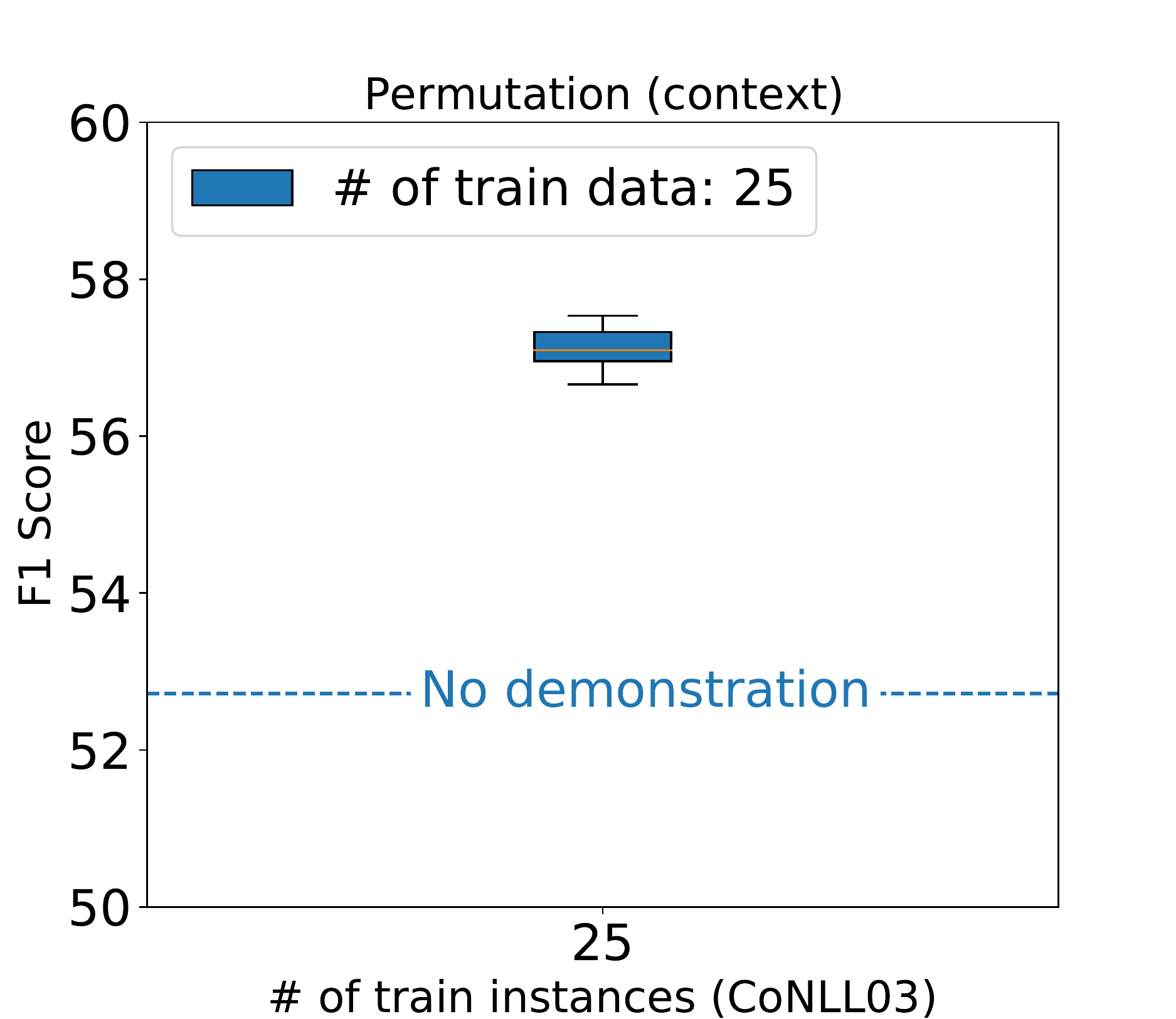}}
  \vspace{-0.1cm}
  \centerline{\small (a) 25 instances}
\end{minipage}
\begin{minipage}{0.23\textwidth}
  \centerline{\includegraphics[width=4.2cm]{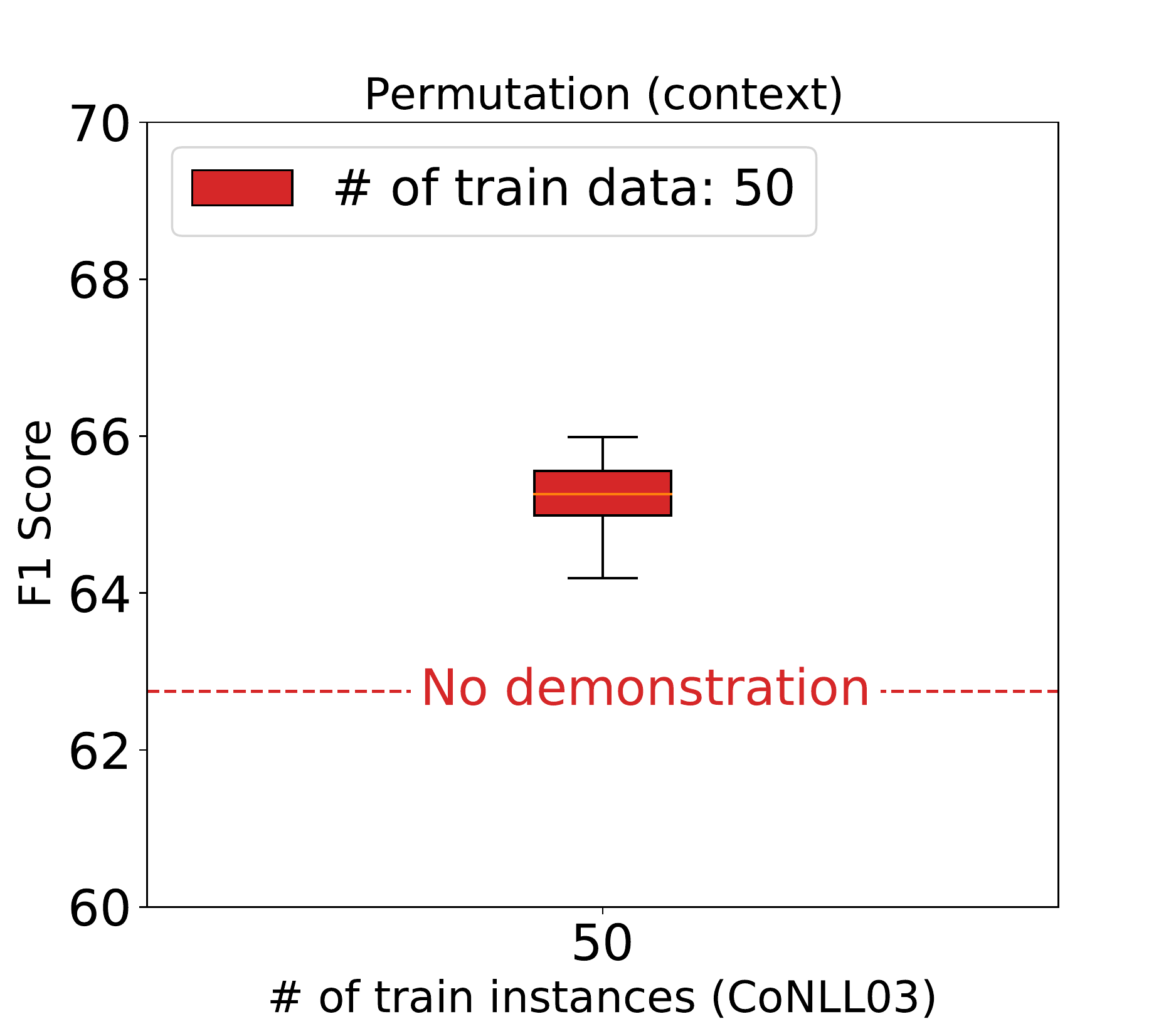}}
  \vspace{-0.1cm}
  \centerline{\small (b) 50 instances}
\end{minipage}
\caption{\textbf{Performance (F1-score) variance by different permutation of entity type orders.} Performance is based on template \texttt{basic}, strategy \texttt{popular}, and CoNLL03.}
\label{fig:permutation}
\end{figure}

\paragraph{Templates of entity-oriented demonstration.}
\textit{entity-oriented demonstration} becomes more effective when not only showing the entity example per each entity type, but also the corresponding instance example as a context. \texttt{context} and \texttt{lexical} consistently outperform \texttt{no-context}.
We explore other templates as well, and these three are the best among them.
We present details on Appendix~\ref{sec:appendix-template}.
To see whether the order of entity type in \textit{entity-oriented demonstration} affects the performance, we present analysis of entity type permutation, e.g., \texttt{person} - \texttt{organization} - \texttt{location} - \texttt{miscellaneous}.
There is no clear pattern of which entity type order is better (spearman correlation between F1-scores over different entity type orders with 25 and 50 training instances < 0), but all the permutations outperform the baseline as shown in Figure~\ref{fig:permutation}, which show that \textit{demonstration-based learning} can be effective regardless of the order (See Appendix Figure~\ref{fig:permutation_appendix}).

\begin{figure}[t]
\vspace{-0.4cm}
\hspace{0.1cm}
\begin{minipage}{0.23\textwidth}
  \centerline{\includegraphics[width=4.2cm]{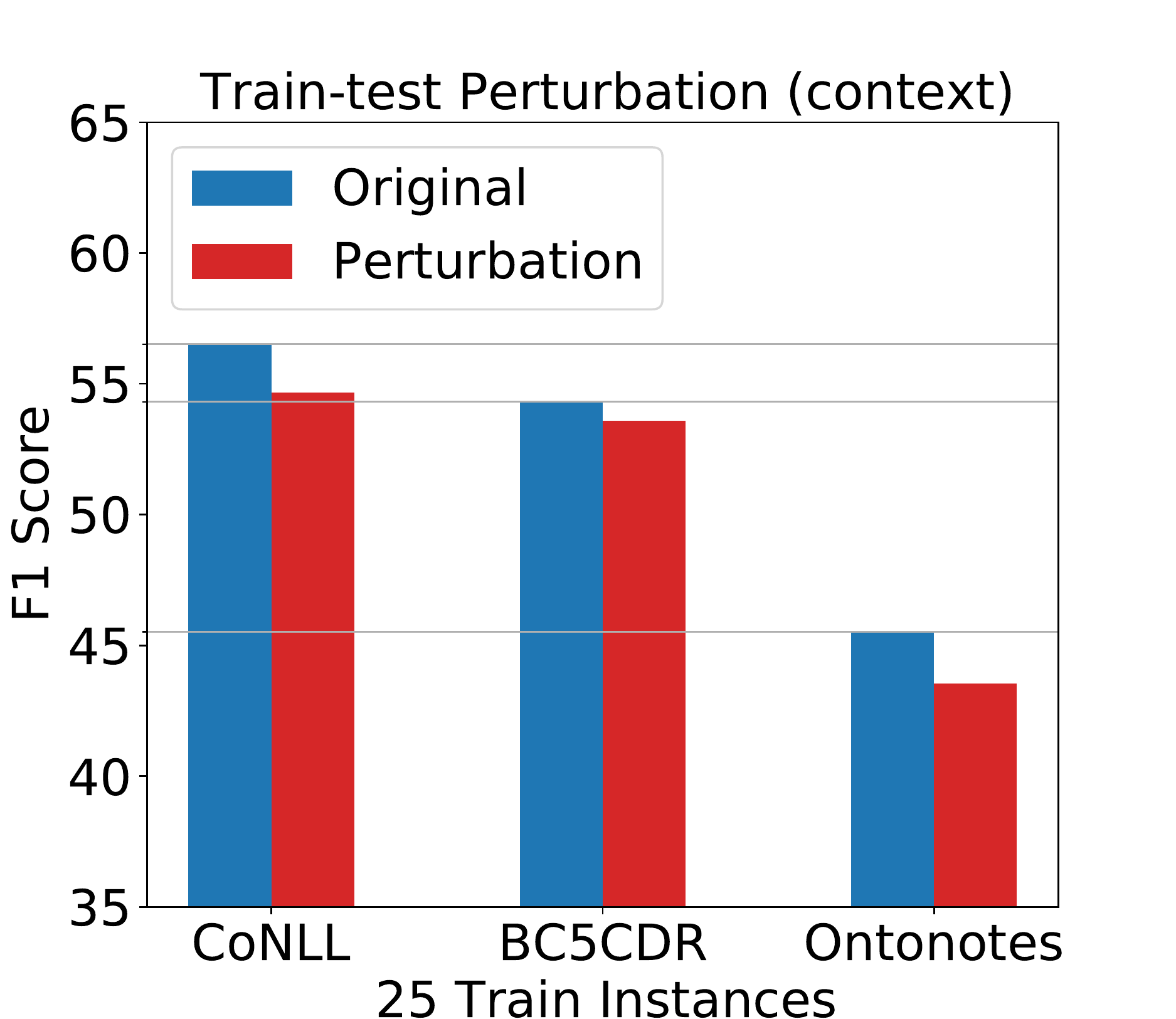}}
  \vspace{-0.1cm}
  \centerline{\small (a) Test perturbation}
\end{minipage}
\begin{minipage}{0.23\textwidth}
  \centerline{\includegraphics[width=4.2cm]{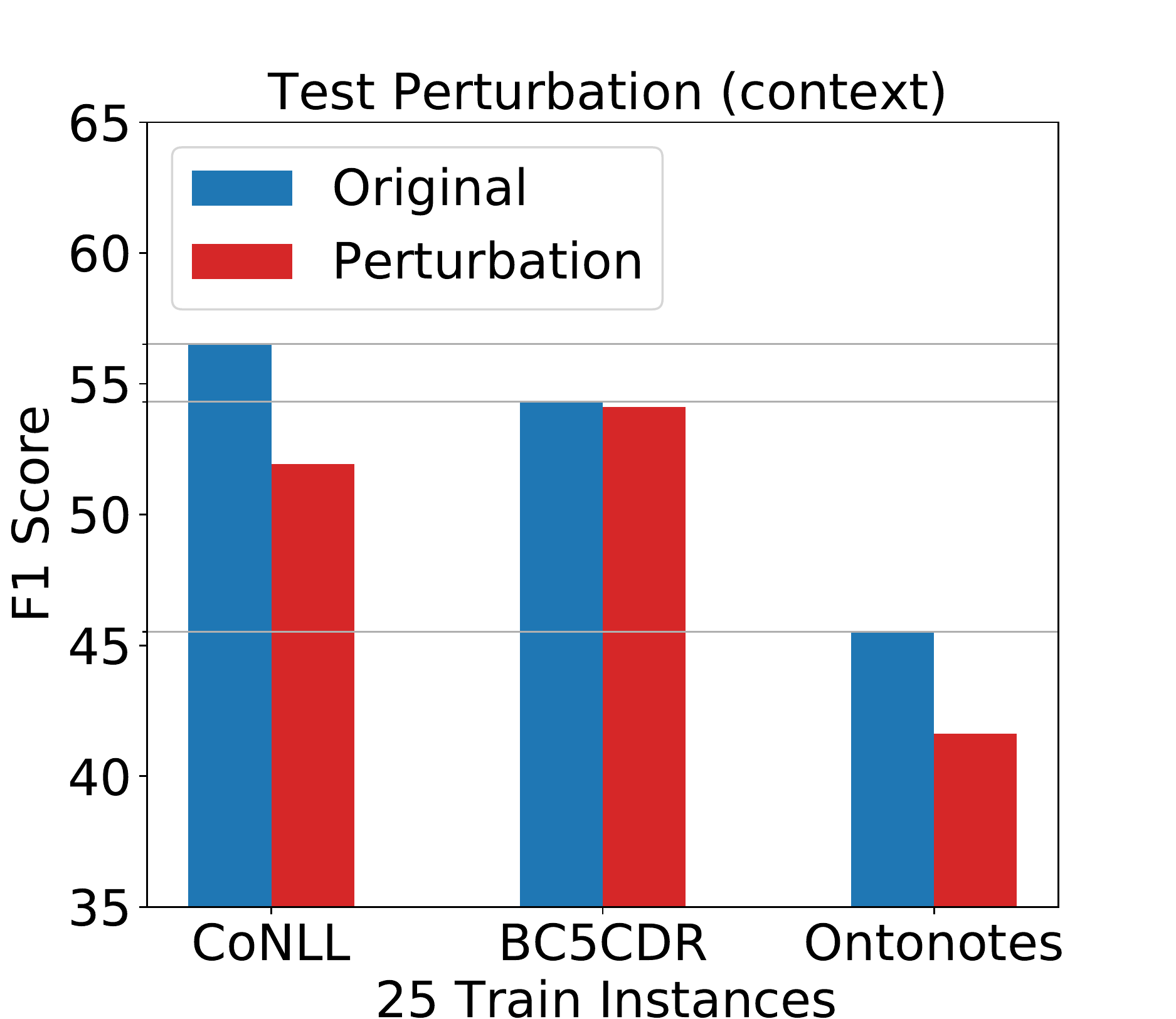}}
  \vspace{-0.1cm}
  \centerline{\small (b) Train-Test perturbation}
\end{minipage}
\caption{\textbf{Performance (F1-score) difference between original and perturbed demonstration.} Performance is based on template \texttt{basic}, strategy \texttt{popular}, and CoNLL03 25 train instances. 
}
\label{fig:perturb}
\end{figure}

\paragraph{Demonstration perturbation.}
To investigate whether the model really learns from demonstration, we explore the performance of our approach with perturbed demonstration which selects random entities, labels, and context sentences as demonstration.
Here, we present two studies: (1) \textit{Test perturbation} which train with correct demonstration and test with perturbed demonstration; and (2) \textit{Train-test perturbation} which both train and test with perturbed demonstration.
Figure~\ref{fig:perturb} shows perturbed demonstration disturbs the model in a large margin for both case.
This shows that the model affects by demonstration, and proper demonstration can improve the model's performance.
Full results are available in Appendix Table~\ref{tab:perturbation_ablation}.

\begin{table}[!t]
	\centering
	\small
	\resizebox{0.5\textwidth}{!}{
		\begin{tabular}{cccccccc}
            \toprule
            \multirow{2}{*}{\textbf{Train}} & \multirow{2}{*}{\textbf{Infer}}  & \multicolumn{2}{c}{\textbf{CoNLL03}} & \multicolumn{2}{c}{\textbf{Ontonotes 5.0}} & \multicolumn{2}{c}{\textbf{BC5CDR}} \\
            \cmidrule(lr){3-4} \cmidrule(lr){5-6} \cmidrule(lr){7-8} & & 
            25 & 50 & 25 & 50 & 25 & 50 \\
            \midrule
            X & X & 52.72 \textsubscript{ $\pm$2.44} & 62.75 \textsubscript{ $\pm$0.98} & 38.97 \textsubscript{ $\pm$4.62} & 54.51 \textsubscript{ $\pm$3.27} & 52.56 \textsubscript{ $\pm$0.46} & 60.20 \textsubscript{ $\pm$2.01} \\
            
            X & O & 51.24 \textsubscript{ $\pm$2.10} & 61.02 \textsubscript{ $\pm$2.05} & 40.48 \textsubscript{ $\pm$3.90} & 52.12 \textsubscript{ $\pm$3.85} & 52.16 \textsubscript{ $\pm$0.55} & 58.12 \textsubscript{ $\pm$1.67}\\

            O & X & 37.71 \textsubscript{ $\pm$4.65} & 53.17 \textsubscript{ $\pm$3.47} & 31.98 \textsubscript{ $\pm$4.25} & 45.27 \textsubscript{ $\pm$5.19} & 51.94 \textsubscript{ $\pm$1.04} & 57.73 \textsubscript{ $\pm$1.52} \\
            
            O & O & 56.52 \textsubscript{ $\pm$3.34} & 64.47 \textsubscript{ $\pm$2.35} & 45.52 \textsubscript{ $\pm$4.69} & 58.40 \textsubscript{ $\pm$3.24} & 54.31 \textsubscript{ $\pm$0.80} & 61.31 \textsubscript{ $\pm$1.51} \\
            \midrule
        \end{tabular}
	}
	\caption{\textbf{Effects of demonstration (F1-score)} with/without the demonstration (denoted by ``O" and ``X", respectively) at training and inference time.
	}
	\label{tab:effect}
\end{table}

\paragraph{Effects of demonstration in train \& inference.}

Table~\ref{tab:effect} shows the effects of demonstration in training and inference stage. A comparison of row 0 with row 3 shows that applying demonstration in the training stage but not in the inference stage would make the model perform worse than the fine-tuning baseline. This is another evidence that consistency of demonstration is essential to the method's performance.

\paragraph{Fully supervised setting.}
Table~\ref{tab:full} shows the performance in fully supervised setting, where the train data is sufficient.
We can see that demonstration-based learning yields similar performance as baselines (p-value < 0.1), which shows that demonstrations are rather redundant when data is abundant.

\begin{table}[!t]
	\centering
	\small
	\resizebox{0.48\textwidth}{!}{
		\begin{tabular}{ccccccc}
            \toprule
            \multirow{2}{*}{\textbf{Strategy}} & \multirow{2}{*}{\textbf{Template}}  & \multicolumn{2}{c}{\textbf{CoNLL03}} & \multicolumn{2}{c}{\textbf{BC5CDR}} \\
            \cmidrule(lr){3-4} \cmidrule(lr){5-6} & & 
            50\% & 100\% & 50\% & 100\% \\
            \midrule
            - & - & 91.24 \textsubscript{ $\pm$0.13} & 91.82 \textsubscript{ $\pm$0.12} & 84.58 \textsubscript{ $\pm$0.17} & 85.89 \textsubscript{ $\pm$0.32} \\
            \texttt{random} & \texttt{context} & 90.60 \textsubscript{ $\pm$0.13} & 91.22 \textsubscript{ $\pm$0.38} & 84.32 \textsubscript{ $\pm$0.07} & 85.58 \textsubscript{ $\pm$0.14} \\
            \texttt{popular} & \texttt{context} & 90.81 \textsubscript{ $\pm$0.11} & 91.85 \textsubscript{ $\pm$0.07} & 84.12 \textsubscript{ $\pm$0.48} & 85.61 \textsubscript{ $\pm$0.12} \\

            \midrule
        \end{tabular}
	}
	\caption{\textbf{Performance (F1-score) in fully supervised setting} by different percentages of train data.
	}
	\label{tab:full}
\end{table}

%% file: section/7_conclusion.tex
\section{Conclusion}
In this paper, we propose \textit{demonstration-based learning} for named entity recognition.
Specifically, we present \textit{entity-oriented demonstration} and \textit{instance-oriented demonstration} and show that they successfully guide the model towards better understandings of the task in low-resource settings.
We observe that \textit{entity-oriented demonstration} is more effective than \textit{instance-oriented demonstration}, and \texttt{search} strategy consistently outperforms all other variants.
Moreover, we find that consistent demonstration for all the instances is crucial to the superior performance of our approach.
We believe that our work provides valuable cost reduction when domain-expert annotations are too expensive and opens up possibilities for future work in automatic demonstration search for few-shot named entity recognition. 

\section*{Acknowledgements}
Dong-Ho Lee, Akshen Kadakia, Mahak Agarwal, Jay Pujara, and Xiang Ren’s work on this project was funded by the Defense Advanced Research Projects Agency with award W911NF-19-20271.
We would like to thank the reviewers for their constructive feedback.

%% file: section/99_appendix.tex
\clearpage
\appendix

\section{Template Analysis}
\label{sec:appendix-template}

Here we present 4 other variants of templates that we have not presented in \textit{entity-oriented demonstration}:
(1) \texttt{context-all} shows selected $e$ per $l$ along with an instance sentence $s$ that contains $e$ as a type of $l$. Unlike \texttt{context}, it shows all the $e \in s$.
For each triple of $(e,l,s)$, it is modified into \text{``$s$. $e_{1}$ \textit{is} $l_{1}$. \ldots $e_{n}$ \textit{is} $l_{n}$."} and concatenated with \textit{[SEP]}. 
(2) \texttt{lexical-all} shows selected $e$ per $l$ in instance example $s$ and further replaces the entity span $e$ by its label string $l$.
Unlike \texttt{lexical}, it replaces all the $e \in s$ by its label string $l$.
(3) \texttt{structure} follows augmented natural language format, which is a structured format~\cite{paolini2021structured}. It shows selected $e$ per $l$ along with an instance sentence $s$ that contains $e$ as a type of $l$. For each triple of $(e,l,s)$, $e$ in $s$ is replaced with \textit{[ $e$ | $l$ ]} and concatenated with \textit{[SEP]}.
(4) \texttt{structure-all} also follows augmented natural language format, and shows selected $e$ per $l$ along with an instance sentence $s$ that contains $e$ as a type of $l$. Unlike \texttt{structure} it shows all the $e \in s$. For each triple of $(e,l,s)$, for each $e_{i}$ in $s$ it is replaced with \textit{[ $e_{i}$ | $l_{i}$ ]} and concatenated with \textit{[SEP]}.done
Table.~\ref{tab:ablation} shows that \texttt{context} and \texttt{lexical} are more effective than others.

\begin{table*}[!t]
	\centering
	\small
	\resizebox{\textwidth}{!}{
		\begin{tabular}{ccccccccccccc}
            \toprule
            \multirow{2}{*}{\textbf{Template}} & \multicolumn{4}{c}{\textbf{CoNLL03}} & \multicolumn{4}{c}{\textbf{Ontonotes 5.0}} & \multicolumn{4}{c}{\textbf{BC5CDR}} \\
            \cmidrule(lr){2-5} \cmidrule(lr){6-9} \cmidrule(lr){10-13} & 
            50 & 100 & 150 & 200 & 50 & 100 & 150 & 200 & 50 & 100 & 150 & 200 \\
            \midrule
            - & 58.51 \textsubscript{$\pm$2.99} & 69.44 \textsubscript{$\pm$4.40} & 73.94 \textsubscript{$\pm$5.69} & 75.83 \textsubscript{$\pm$5.61} & 46.34 \textsubscript{$\pm$4.46} & 60.36 \textsubscript{$\pm$7.52} & 65.69 \textsubscript{$\pm$7.41} & 68.81 \textsubscript{$\pm$7.52} & 55.68 \textsubscript{$\pm$5.33} & 64.24 \textsubscript{$\pm$2.79} & 68.37 \textsubscript{$\pm$2.55} & 71.09 \textsubscript{$\pm$2.84} \\
            \midrule
            \texttt{no-context} & 58.23 \textsubscript{$\pm$3.09} & 69.52 \textsubscript{$\pm$3.32} & 72.99 \textsubscript{$\pm$4.63} & 76.33 \textsubscript{$\pm$4.49} & 49.63 \textsubscript{$\pm$3.49} & 62.10 \textsubscript{$\pm$6.53} & 67.48 \textsubscript{$\pm$6.20} & 69.68 \textsubscript{$\pm$7.00} & 56.04 \textsubscript{$\pm$5.34} & 64.32 \textsubscript{$\pm$2.63} & \underline{68.55 \textsubscript{$\pm$2.82}} & 71.14 \textsubscript{$\pm$3.29} \\
            \texttt{context} & 59.14 \textsubscript{$\pm$2.53} & \bf 69.75 \textsubscript{$\pm$3.50} & 73.35 \textsubscript{$\pm$4.24} & 76.59 \textsubscript{$\pm$3.96} & 52.93 \textsubscript{$\pm$4.64} & \underline{63.37 \textsubscript{$\pm$7.02}} & 68.05 \textsubscript{$\pm$6.40} & \bf 70.23 \textsubscript{$\pm$6.28} & 57.10 \textsubscript{$\pm$4.55} & 64.42 \textsubscript{$\pm$3.14} & 68.46 \textsubscript{$\pm$2.94} & 71.27 \textsubscript{$\pm$3.43} \\
            \texttt{lexical} & \underline{59.62 \textsubscript{$\pm$3.12}} & 69.22 \textsubscript{$\pm$3.94} & \bf 74.23 \textsubscript{$\pm$4.26} & \underline{77.01 \textsubscript{$\pm$4.07}} & 52.69 \textsubscript{$\pm$4.47} & 62.80 \textsubscript{$\pm$7.12} & 67.78 \textsubscript{$\pm$6.02} & 70.02 \textsubscript{$\pm$6.86} & \underline{57.83 \textsubscript{$\pm$4.53}} & 64.52 \textsubscript{$\pm$3.36} & 68.51 \textsubscript{$\pm$2.57} & 71.14 \textsubscript{$\pm$3.04} \\
            \texttt{structure} & \bf 60.61 \textsubscript{$\pm$2.60} & 68.35 \textsubscript{$\pm$3.85} & 73.95 \textsubscript{$\pm$4.60} & 76.56 \textsubscript{$\pm$4.38} & \bf 53.35 \textsubscript{$\pm$3.59} & \bf 63.45 \textsubscript{$\pm$6.23} & \underline{68.10 \textsubscript{$\pm$5.99}} & 69.99 \textsubscript{$\pm$6.74} & 57.45 \textsubscript{$\pm$4.79} & \bf 64.72 \textsubscript{$\pm$2.79} & 68.32 \textsubscript{$\pm$2.77} & \bf 71.55 \textsubscript{$\pm$3.20} \\
            \texttt{context-all} & 58.82 \textsubscript{$\pm$2.01} & 69.22 \textsubscript{$\pm$3.37} & 71.22 \textsubscript{$\pm$3.45} & 76.07 \textsubscript{$\pm$4.53} & 52.85 \textsubscript{$\pm$4.23} & 62.80 \textsubscript{$\pm$7.40} & \bf 68.22 \textsubscript{$\pm$6.18} & 69.87 \textsubscript{$\pm$6.63} & \bf 57.92 \textsubscript{$\pm$4.58} & \underline{64.69 \textsubscript{$\pm$2.72}} & \bf 68.83 \textsubscript{$\pm$2.28} & \underline{71.32 \textsubscript{$\pm$3.13}} \\
            \texttt{lexical-all} & 59.34 \textsubscript{$\pm$2.72} & \underline{69.71 \textsubscript{$\pm$3.65}} & \underline{74.16 \textsubscript{$\pm$4.47}} & \bf 77.31 \textsubscript{$\pm$4.04} & 52.46 \textsubscript{$\pm$4.47} & 63.03 \textsubscript{$\pm$7.33} & 67.22 \textsubscript{$\pm$6.82} & \underline{70.21 \textsubscript{$\pm$6.68}} & 56.76 \textsubscript{$\pm$5.01} & 64.42 \textsubscript{$\pm$2.91} & 68.05 \textsubscript{$\pm$3.18} & 71.17 \textsubscript{$\pm$3.13} \\
            \texttt{structure-all} & 59.27 \textsubscript{$\pm$2.28} & 69.17 \textsubscript{$\pm$3.28} & 73.69 \textsubscript{$\pm$4.43} & 76.14 \textsubscript{$\pm$4.21} & \underline{53.33 \textsubscript{$\pm$4.39}} & 62.69 \textsubscript{$\pm$6.48} & 67.99 \textsubscript{$\pm$6.08} & 70.09 \textsubscript{$\pm$6.34} & 56.99 \textsubscript{$\pm$5.56} & 64.42 \textsubscript{$\pm$2.71} & 68.43 \textsubscript{$\pm$2.94} & 70.92 \textsubscript{$\pm$3.12} \\
            \midrule
        \end{tabular}
	}
	\caption{\textbf{Template performance comparison (F1-score) in \texttt{popular} strategy} on CoNLL03, Ontonotes 5.0, and BC5CDR by different number of training instances. We randomly sample $k$ training instances with a constraint that sampled instances should cover all the IOBES labels in the whole dataset. Best variants are \textbf{bold} and second best ones are \underline{underlined}. For efficient training, here the batch size is 10.}
	\label{tab:ablation}
\end{table*}

\section{Effects of Batch Size}
\label{sec:appendix-batch}
Table~\ref{tab:indomain_ablation} shows the main results in Table~\ref{tab:indomain} with batch size 10.
Overall performance is much lower than Table~\ref{tab:indomain}.
It shows that choosing a lower batch size is important in a extremely low resource, where the number of train data is 25 or 50.

\begin{table*}[!t]
	\centering
	\small
	\resizebox{\textwidth}{!}{
		\begin{tabular}{lcccccccc}
            \toprule
            \multirow{2}{*}{\textbf{Demonstration}} & \multirow{2}{*}{\textbf{Strategy}} & 
            \multirow{2}{*}{\textbf{Template}} & \multicolumn{2}{c}{\textbf{CoNLL03}} & \multicolumn{2}{c}{\textbf{Ontonotes 5.0}} & \multicolumn{2}{c}{\textbf{BC5CDR}} \\
            \cmidrule(lr){4-5} \cmidrule(lr){6-7} \cmidrule(lr){8-9} & & &
            25 & 50 & 25 & 50 & 25 & 50 \\
            \midrule
            No Demonstration & - & - & 42.65 \textsubscript{$\pm$4.77} & 60.14 \textsubscript{$\pm$3.28} & 29.11 \textsubscript{$\pm$5.21} & 49.00 \textsubscript{$\pm$4.92} & 50.59 \textsubscript{$\pm$3.64} & 57.44 \textsubscript{$\pm$4.51} \\
            \midrule
            Instance-oriented  
            & \texttt{SBERT} & \texttt{lexical} & 39.25 \textsubscript{$\pm$5.57} & 54.13 \textsubscript{$\pm$4.72} & 26.41 \textsubscript{$\pm$5.84} & 41.09 \textsubscript{$\pm$4.07} & 47.08 \textsubscript{$\pm$5.65} & 50.78 \textsubscript{$\pm$4.77} \\
            Demonstration & (\texttt{variable}) & \texttt{context} & 41.09 \textsubscript{$\pm$5.82} & 59.92 \textsubscript{$\pm$4.78} & 30.55 \textsubscript{$\pm$6.61} & 48.46 \textsubscript{$\pm$5.03} & 51.72 \textsubscript{$\pm$5.81} & 57.53 \textsubscript{$\pm$4.58} \\
            \cmidrule(lr){2-9}
            & \texttt{BERTScore} & \texttt{lexical} & 40.27 \textsubscript{$\pm$6.36} & 55.85 \textsubscript{$\pm$4.39} & 23.84 \textsubscript{$\pm$6.10} & 41.34 \textsubscript{$\pm$3.99} & 47.24 \textsubscript{$\pm$5.53} & 49.73 \textsubscript{$\pm$5.43} \\
            & (\texttt{variable}) & \texttt{context} & 41.42 \textsubscript{$\pm$6.5} & 60.65 \textsubscript{$\pm$4.64} & 25.79 \textsubscript{$\pm$5.74} & 42.21 \textsubscript{$\pm$3.23} & 51.85 \textsubscript{$\pm$5.87} & 56.68 \textsubscript{$\pm$5.31} \\
            \midrule
            Entity-oriented 
            & \texttt{random} & \texttt{no-context} & 44.19 \textsubscript{$\pm$4.98} & 58.87 \textsubscript{$\pm$3.80} & 33.07 \textsubscript{$\pm$7.14} & 50.02 \textsubscript{$\pm$5.48} & 51.07 \textsubscript{$\pm$2.85} & 58.08 \textsubscript{$\pm$3.45}  \\
            Demonstration & (\texttt{variable}) & \texttt{lexical} & 46.83 \textsubscript{$\pm$3.69} & 59.94 \textsubscript{$\pm$3.82} & 34.52 \textsubscript{$\pm$6.58} & 50.69 \textsubscript{$\pm$5.64} & 51.72 \textsubscript{$\pm$2.75} & 57.62 \textsubscript{$\pm$3.33} \\
            & & \texttt{context} & 47.39 \textsubscript{$\pm$3.89} & 59.81 \textsubscript{$\pm$3.58} & 35.39 \textsubscript{$\pm$7.10} & 50.80 \textsubscript{$\pm$5.63} & 51.86 \textsubscript{$\pm$2.71} & 58.12 \textsubscript{$\pm$2.97} \\
            \cmidrule(lr){2-9}
            & \texttt{popular} & \texttt{no-context} & 46.51 \textsubscript{$\pm$4.50} & 60.67 \textsubscript{$\pm$2.97} & 34.50 \textsubscript{$\pm$6.51} & 52.38 \textsubscript{$\pm$4.61} & 51.12 \textsubscript{$\pm$3.28} & 57.71 \textsubscript{$\pm$4.46} \\
            & (\texttt{fixed}) & \texttt{lexical} & 49.92 \textsubscript{$\pm$3.52} & 60.75 \textsubscript{$\pm$3.29} & 36.99 \textsubscript{$\pm$6.11} & 54.56 \textsubscript{$\pm$4.59} & 52.23 \textsubscript{$\pm$3.56} & 58.53 \textsubscript{$\pm$4.64} \\
            & & \texttt{context} & 50.54 \textsubscript{$\pm$3.43} & 61.08 \textsubscript{$\pm$3.10} & \underline{37.97 \textsubscript{$\pm$6.14}} & \underline{54.66 \textsubscript{$\pm$4.43}} & 52.78 \textsubscript{$\pm$2.71} & 58.69 \textsubscript{$\pm$4.17} \\
            \cmidrule(lr){2-9}
            & \texttt{search} & \texttt{no-context} & 47.80 \textsubscript{$\pm$3.45} & 60.74 \textsubscript{$\pm$3.50} & 34.44 \textsubscript{$\pm$6.04} & 53.06 \textsubscript{$\pm$4.78} & 51.65 \textsubscript{$\pm$2.94} & 58.32 \textsubscript{$\pm$4.08}  \\
            & (\texttt{fixed}) & \texttt{lexical} & \underline{50.77 \textsubscript{$\pm$3.32}} & \underline{61.67 \textsubscript{$\pm$3.66}} & 37.41 \textsubscript{$\pm$6.74} & 54.62 \textsubscript{$\pm$4.17} & \underline{52.89 \textsubscript{$\pm$3.43}} & \underline{58.80 \textsubscript{$\pm$4.23}} \\
            & & \texttt{context} & \bf 51.57 \textsubscript{$\pm$3.25} & \bf 62.26 \textsubscript{$\pm$2.75} & \bf 38.17 \textsubscript{$\pm$6.60} & \bf 54.99 \textsubscript{$\pm$4.09} & \bf 53.01 \textsubscript{$\pm$3.42} & \bf 59.15 \textsubscript{$\pm$3.96} \\
            \midrule
        \end{tabular}
	}
	\caption{\textbf{In-domain performance comparison (F1-score)} on CoNLL03, Ontonotes 5.0, and BC5CDR by different number of training instances. We randomly sample $k$ training instances with a constraint that sampled instances should cover all the IOBES labels in the whole dataset. Best variants are \textbf{bold} and second best ones are \underline{underlined}.
	Scores are average of 15 runs (5 different sub-samples and 3 random seeds) and the backbone LM model is \texttt{bert-base-cased}. Unlike Table~\ref{tab:indomain}, here the batch size is 10.
	}
	\label{tab:indomain_ablation}
\end{table*}

\begin{figure*}[t]
\hspace{0.1cm}
\centerline{\includegraphics[width=15cm]{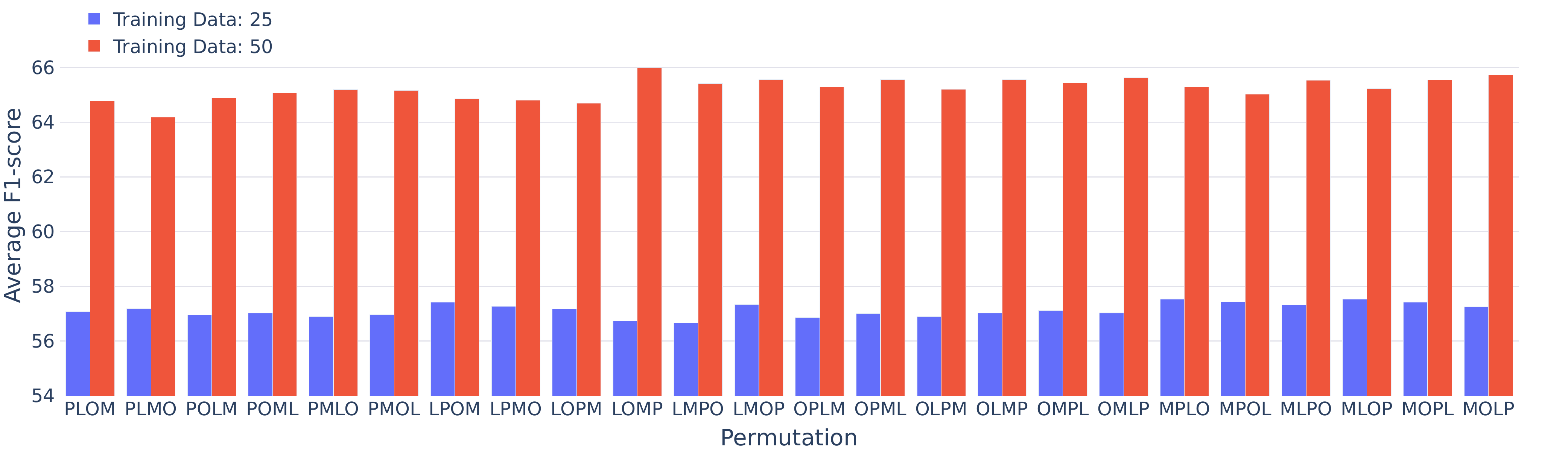}}
\caption{\textbf{Performance comparison (F1-score) by different entity type order in entity-oriented demonstration.} Performance is based on template \texttt{basic} and strategy \texttt{popular}, and dataset is CoNLL03.
We construct the demonstration by different entity type order (P: Person, L: Location, O: Organization, M: Miscellaneous). Scores are average of 15 runs (5 different subsamples and 3 random seeds).
}
\label{fig:permutation_appendix}
\end{figure*}

\begin{table*}[!t]
	\centering
	\small
	\resizebox{\textwidth}{!}{
		\begin{tabular}{cccccccc}
            \toprule
            \multirow{2}{*}{\textbf{Template}} & 
            \multirow{2}{*}{\textbf{Test}} & \multicolumn{2}{c}{\textbf{CoNLL03}} & \multicolumn{2}{c}{\textbf{Ontonotes 5.0}} & \multicolumn{2}{c}{\textbf{BC5CDR}} \\
            \cmidrule(lr){3-4} \cmidrule(lr){5-6} \cmidrule(lr){7-8} & \textbf{Perturbation} &
            25 & 50 & 25 & 50 & 25 & 50 \\
            \midrule
            \texttt{no-context} & X & 54.34 \textsubscript{$\pm$3.33} & 64.30 \textsubscript{$\pm$2.76} & 43.02 \textsubscript{$\pm$4.33} & 56.65 \textsubscript{$\pm$3.35} & 53.86 \textsubscript{$\pm$0.86} & 60.51 \textsubscript{$\pm$1.77} \\
            \texttt{no-context} & O & 53.83 \textsubscript{ $\pm$3.65} & 62.86 \textsubscript{ $\pm$2.16} & 41.59 \textsubscript{ $\pm$5.76} & 54.63 \textsubscript{ $\pm$3.89} & 53.06 \textsubscript{ $\pm$0.84} & 59.67 
            \textsubscript{ $\pm$1.55}\\
            \midrule
            \texttt{context} & X & 56.52 \textsubscript{ $\pm$3.34} & 64.47 \textsubscript{ $\pm$2.35} & 45.52 \textsubscript{ $\pm$4.69} & 58.40 \textsubscript{ $\pm$3.24} & 54.31 \textsubscript{ $\pm$0.8} & 61.31 
            \textsubscript{ $\pm$1.51} \\
            \texttt{context} & O & 51.93 \textsubscript{ $\pm$5.96} & 62.21 \textsubscript{ $\pm$2.66} & 41.63 \textsubscript{ $\pm$5.61} & 53.80 \textsubscript{ $\pm$4.74} & 54.12 \textsubscript{ $\pm$0.95} & 59.63 
            \textsubscript{ $\pm$1.24} \\
            \midrule
        \end{tabular}
	}
	\resizebox{\textwidth}{!}{
		\begin{tabular}{cccccccc}
            \toprule
            \multirow{2}{*}{\textbf{Template}} & 
            \multirow{2}{*}{\textbf{Train-Test}} & \multicolumn{2}{c}{\textbf{CoNLL03}} & \multicolumn{2}{c}{\textbf{Ontonotes 5.0}} & \multicolumn{2}{c}{\textbf{BC5CDR}} \\
            \cmidrule(lr){3-4} \cmidrule(lr){5-6} \cmidrule(lr){7-8} & \textbf{Perturbation} &
            25 & 50 & 25 & 50 & 25 & 50 \\
            \midrule
            \texttt{no-context} & X & 54.34 \textsubscript{$\pm$3.33} & 64.30 \textsubscript{$\pm$2.76} & 43.02 \textsubscript{$\pm$4.33} & 56.65 \textsubscript{$\pm$3.35} & 53.86 \textsubscript{$\pm$0.86} & 60.51 \textsubscript{$\pm$1.77} \\
            \texttt{no-context} & O & 54.13 \textsubscript{ $\pm$2.31} & 62.88 \textsubscript{ $\pm$2.36} & 42.34 \textsubscript{ $\pm$4.91} & 55.17 \textsubscript{ $\pm$3.46} & 53.16 \textsubscript{ $\pm$0.70} & 59.93 
            \textsubscript{ $\pm$2.31}\\
            \midrule
            \texttt{context} & X & 56.52 \textsubscript{ $\pm$3.34} & 64.47 \textsubscript{ $\pm$2.35} & 45.52 \textsubscript{ $\pm$4.69} & 58.40 \textsubscript{ $\pm$3.24} & 54.31 \textsubscript{ $\pm$0.8} & 61.31 
            \textsubscript{ $\pm$1.51} \\
            \texttt{context} & O & 54.67 \textsubscript{ $\pm$3.04} & 63.93 \textsubscript{ $\pm$1.92} & 43.55 \textsubscript{ $\pm$5.64} & 56.09 \textsubscript{ $\pm$3.37} & 53.59 \textsubscript{ $\pm$0.82} & 59.45 
            \textsubscript{ $\pm$1.66} \\
            \midrule
        \end{tabular}
	}
	\caption{\textbf{Perturbation Analysis.}}
	\label{tab:perturbation_ablation}
\end{table*}